\documentclass{article}
\pdftrailerid{}
\usepackage{iclr2027_conference,times}
  \iclrfinalcopy
\usepackage[utf8]{inputenc}
\usepackage[T1]{fontenc}
\usepackage{amsmath,amssymb,graphicx,booktabs,array,url,xcolor,pifont,microtype}
\usepackage{colortbl}
\usepackage{tabularx}
\usepackage{multirow}
\usepackage{wrapfig}
\usepackage{enumitem}
\usepackage{etoolbox}
\AtBeginEnvironment{table}{\setlength{\belowcaptionskip}{4pt}}
\AtBeginEnvironment{wraptable}{\setlength{\belowcaptionskip}{4pt}\setlength{\abovecaptionskip}{0pt}}
\definecolor{ourblue}{rgb}{0.368,0.507,0.71}
\definecolor{groupyellow}{rgb}{1.0,0.976,0.86}
\newcommand{\grouprow}[2]{\rowcolor{groupyellow}[0pt][0pt]\multicolumn{#1}{@{}c@{}}{\textit{#2}\strut}}
\usepackage{hyperref}
\hypersetup{colorlinks, linkcolor=ourblue, citecolor=ourblue, urlcolor=ourblue}

\DeclareUnicodeCharacter{00B7}{\ensuremath{\cdot}}
\DeclareUnicodeCharacter{00A7}{\S}
\DeclareUnicodeCharacter{2013}{--}
\DeclareUnicodeCharacter{2014}{---}
\DeclareUnicodeCharacter{2248}{\ensuremath{\approx}}
\DeclareUnicodeCharacter{00D7}{\ensuremath{\times}}

\newcommand{\cmark}{\ding{51}}
\newcommand{\xmark}{\ding{55}}

\graphicspath{{figures/}}

  \title{$\vcenter{\hbox{\includegraphics[height=2.9em]{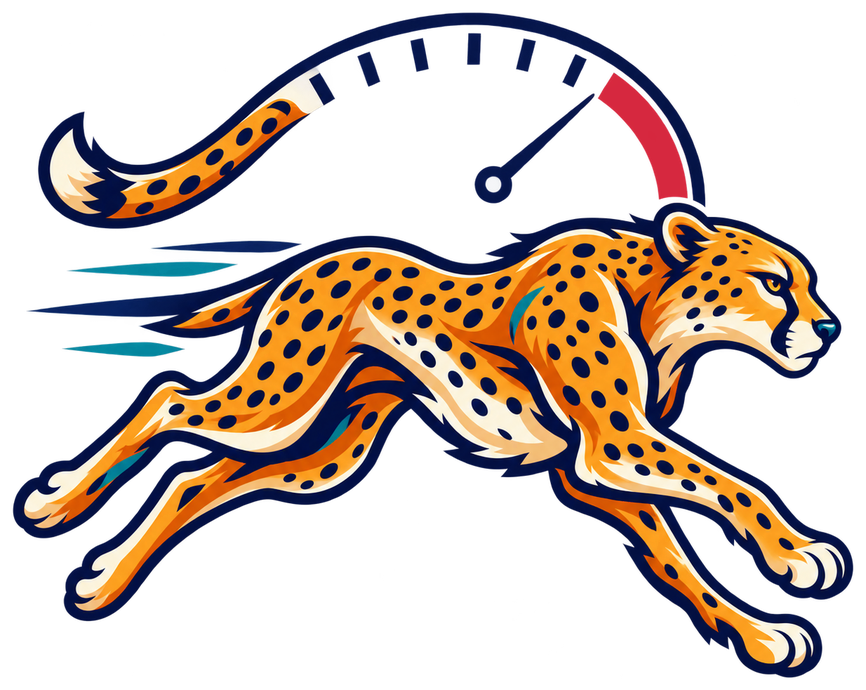}}}$\hspace{0.45em}$\vcenter{\hbox{\begin{tabular}{@{}l@{}}Faster Block-Diffusion Serving with\\Distribution-Free Risk Guarantees\end{tabular}}}$\vspace{14pt}}

\author{%
  \vspace*{4pt}
  \makebox[5.28in][c]{%
  \begin{tabular}{@{}c@{\hspace{0.02in}}c@{\hspace{0.02in}}c@{}}
    \parbox[c]{1.76in}{\centering Jungseob Lee$^{1}$\\[1pt] \fontsize{8pt}{7.5pt}\texttt{omanma1928@korea.ac.kr}} &\vspace*{4pt}
    \parbox[c]{1.76in}{\centering Dongyub Jude Lee$^{2}$\\[1pt] \fontsize{8pt}{7.5pt}\texttt{jude.lee@zoom.us}} &\vspace*{4pt}
    \parbox[c]{1.76in}{\centering Chanjun Park$^{3}$\\[1pt] \fontsize{8pt}{7.5pt}\texttt{chanjun.park@ssu.ac.kr}}\\[4pt]\vspace*{4pt}
    \makebox[0pt][l]{\parbox[c]{1.76in}{\centering Sugyeong Eo$^{4,\dagger}$\\[1pt] \fontsize{8pt}{7.5pt}\texttt{s.eo@yonsei.ac.kr}}} &
    \makebox[0pt][l]{\parbox[c]{1.76in}{\centering Heuiseok Lim$^{1,\dagger}$\\[1pt] \fontsize{8pt}{7.5pt}\texttt{limhseok@korea.ac.kr}}} &
    \makebox[0pt][l]{}
  \end{tabular}%
  }\\[7pt]
  \small
  \makebox[5.28in][c]{%
    \begin{tabular}{c}
      $^{1}$Korea University \hspace{0.1in} $^{2}$Zoom Communications \hspace{0.1in} $^{3}$Soongsil University \hspace{0.1in} $^{4}$Yonsei University Mirae Campus
    \end{tabular}%
  }%
}

\begin{document}
\maketitle
  \begingroup
  \renewcommand{\thefootnote}{\textdagger}
  \footnotetext{Corresponding authors.}
  \endgroup
  \lhead{Preprint}
\suppressfloats[t]  %

\begin{abstract}
Block-diffusion language models are served at hand-picked operating points, such as acceptance thresholds, buffer depth, schedule, checkpoint and precision, and each point is chosen by its mean benchmark accuracy. However, a mean does not tell an operator how often a faster configuration fails on prompts that the slower one answers correctly. On the serving engine and its decode traces, the default commit rule already commits every fully resolved block, a static skip rule captures nearly all of the compute that allocation can save, and self-distillation on engine-decoded targets adds speed at unchanged accuracy. Larger speedups come from lower thresholds, which commit tokens that are still uncertain. We therefore present Redline, a finite-sample procedure that selects operating points, hand-picked or learned, from the correctness of their answers on calibration prompts. Redline keeps the reference-relative risk, the joint probability that the reference answers correctly and a candidate configuration does not, within a user-chosen budget with high probability, and deploys the fastest configuration that passes. It speeds up math at a smaller risk budget than code in both model families, and at a budget of ten percent it deploys a LLaDA2 math configuration that commits over a third more tokens in each forward. It also applies without modification to the acceptance rule of speculative decoding and to weight quantization. On the same calibration data, Redline stays within its stated failure probability, whereas each tolerance of a mean-accuracy rule either gains less speed for some model and task or exceeds the risk budget far more often for another. Code is available at \url{https://github.com/js-lee-AI/Redline}.
\end{abstract}

\section{Introduction}
\label{sec:intro}

\begin{figure}[t]
\centering
\includegraphics[width=\linewidth]{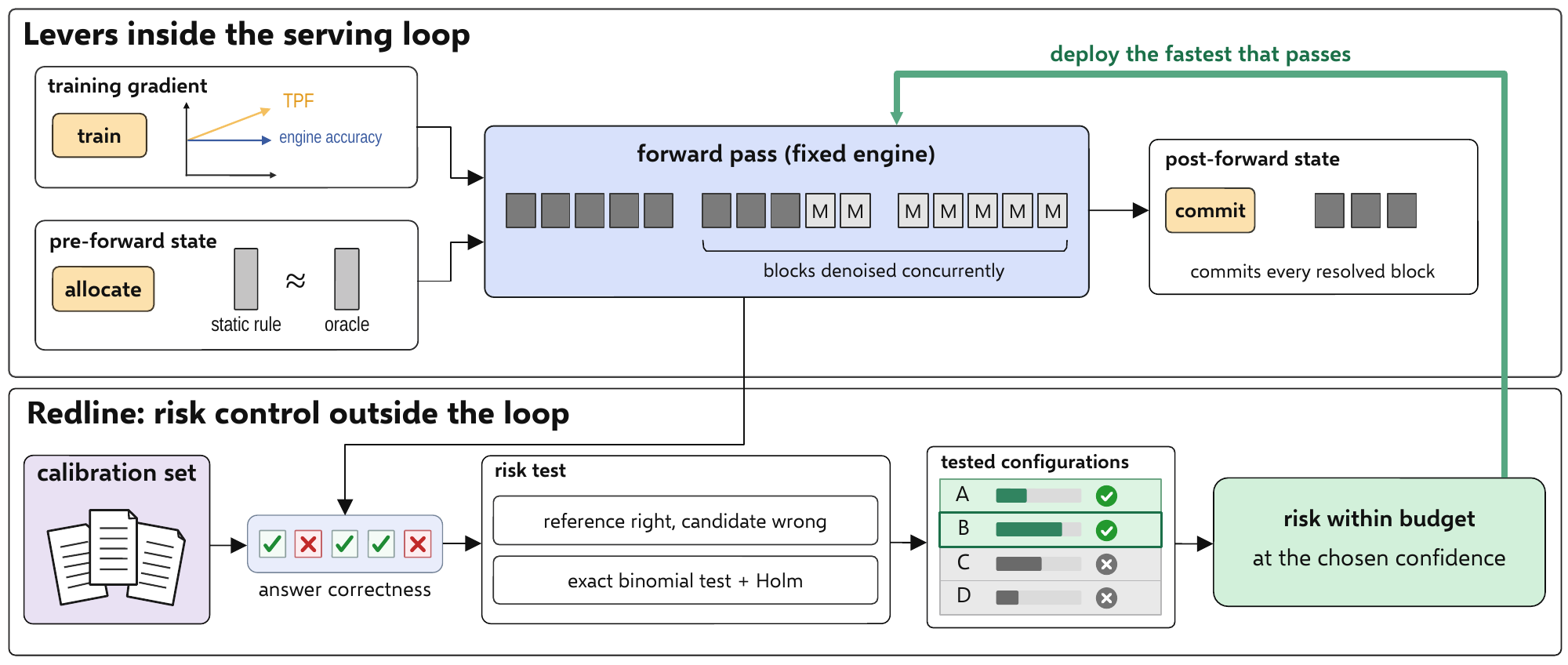}
\caption{Levers inside the serving loop and risk control outside it. Top, the three levers at the decision sites they act on. Bottom, Redline tests every configuration on the correctness of its answers, keeps those that pass (green) and deploys the fastest.}
\label{fig:teaser}
\end{figure}

Block-diffusion language models (BD-LMs) generate text block by block. Each block is produced autoregressively and cached like an autoregressive prefix, while the tokens inside the block are denoised in parallel \citep{arriola2025block,cheng2025sdar,bie2025llada2}. Multi-block engines go further and keep several blocks in flight, which raises the number of tokens committed in each model forward (TPF) at equal or better accuracy \citep{jin2026mbd}. In practice, each such system is released with hand-picked serving settings, and its published operating points are chosen by mean benchmark accuracy \citep{bie2025llada2,bie2026llada21,cheng2025sdar}. A mean does not tell an operator how often, and with what confidence, a faster configuration fails on a prompt that the slower one answers correctly, because the prompts it newly gets right offset those it newly gets wrong.

The default response of the field is to \emph{learn} a better frontier \citep{chen2025dparallel,qian2026d3llm,wang2025d2f,hu2026lightningrl,lee2026acoer,chen2026dmax,bie2026llada21}. None of these methods states a finite-sample bound on deployed quality. We measure how many blocks the engine keeps in flight and the three levers of the serving loop (training, allocation and commitment) on the engine and its decode traces, each against its control. The engine keeps only about two blocks in flight because its deeper blocks wait for text that is not yet generated. The default commit rule already commits every fully resolved block once the blocks before it are committed, and a static skip rule saves nearly all of the compute that allocation can save. Self-distillation on targets that the engine itself decodes gains 5 to 6\% TPF at unchanged accuracy. Larger gains come from lower thresholds, which let the engine commit positions that are still uncertain and condition later blocks on them. A learned checkpoint thus becomes one more candidate for the serving grid, and its risk, like that of a hand-picked setting, shows only in the answers.

Whether an answer is correct can be checked on held-out prompts after serving, and these outcomes suffice to control the risk of an operating point from outside the loop, as Figure~\ref{fig:teaser} shows. For each candidate configuration, we define its reference-relative risk as the \emph{joint} probability, over prompts, that a reference configuration answers correctly and the candidate does not. We present Redline, a procedure that tests every configuration of a searched grid against a user-chosen risk budget with an exact binomial test and a multiplicity correction, and deploys the fastest configuration that passes. The risk of the deployed configuration then stays within the budget at a user-chosen confidence level, in finite samples and without distributional assumptions. Because the risk counts only the prompts that the candidate newly fails, which a user switched to it meets as regressions, it also bounds the net accuracy drop from above. Redline needs no gradient update and no architectural change. Prior distribution-free risk control in decoding calibrates a single early-exit threshold \citep{angelopoulos2021learn,schuster2022confident}. We instead select from a searched grid over several serving settings, across which risk is not monotone.

We apply Redline to two BD-LM families on math and code. In both families, math gains speed at a smaller risk budget than code, and at every budget at which code gains speed, math does too. At a risk budget of 0.10, for instance, the deployed LLaDA2 math configuration commits over a third more tokens in each forward than the reference. Added to the SDAR math grid on MATH-500, the distilled checkpoint is deployed at every budget from 0.18 on, with a TPF gain of 46.7\% against 39.5\% for the best hand-picked configuration.

On the same prompts and at a budget of 0.10, a mean-accuracy rule, which deploys the fastest configuration whose accuracy is within a tolerance of the reference's, exceeds the risk budget in most data splits on SDAR math whenever its tolerance keeps pace with Redline on LLaDA2 math, and so does bounding the net accuracy drop instead. In contrast, Redline stays within its stated failure probability under both the held-out and the pooled measure of risk. Finally, because Redline uses only the correctness of answers and a cost measure, it applies unchanged to the acceptance rule of speculative decoding and to weight quantization, two lossy accelerations of autoregressive serving.

Our contributions are as follows.
\begin{itemize}[leftmargin=*]
\item We introduce Redline, a distribution-free, finite-sample procedure that deploys the fastest configuration whose reference-relative risk it can keep within a user-chosen budget at a user-chosen confidence.
\item We apply it to two block-diffusion families on math and code, show on the same data that mean accuracy hides the regressions that Redline bounds, and transfer it unchanged to speculative decoding and weight quantization.
\item We measure the three in-loop serving levers, each against its control, and find that the large TPF gains depend on lower thresholds, while a checkpoint distilled in the loop adds 5 to 6\% TPF at unchanged accuracy and is tested by Redline like any hand-picked configuration.
\end{itemize}

\section{Background and Setup}
\label{sec:prelim}

\subsection{Block-diffusion LMs and multi-block serving}
\label{sec:prelim-bdlm}

Block-diffusion LMs factorize a sequence into fixed-size blocks that are generated autoregressively, while the tokens inside each block are denoised in parallel under a block-causal mask \citep{arriola2025block}, and the block boundary lets generated blocks be cached like an autoregressive prefix. The serving stack of \citet{jin2026mbd} extends this to a multi-block regime and raises TPF by up to ${\sim}78\%$ at equal or better accuracy. Several blocks are \emph{in flight} at once, denoised concurrently inside a fixed-slot \emph{block buffer}. The blocks in flight are governed by the admission threshold $\tau_{\text{add}}$, the semi-completion threshold $\tau_{\text{semi}}$, which lets later blocks condition on the top-1 context of a predecessor before it is cached, and the acceptance threshold $\tau_{\text{acc}}$, together with the stability and revision thresholds of the engine. We keep the engine fixed throughout.

We measure parallelism by TPF, the number of tokens committed in each model forward. TPF is a \emph{count ratio}, so it does not depend on host speed, kernel choice or batching policy. Its speculative-decoding analogue is the number of accepted tokens for each target forward. In the serving logs of our grids, TPF orders the configurations on each host exactly as measured end-to-end throughput does, with a Spearman correlation of $1.0$, and Appendix~\ref{app:tpf-tps} reports these measurements.

\subsection{Distribution-free risk control}
\label{sec:prelim-dfrc}

A bounded prompt-level $0/1$ loss induces a \emph{risk function} $R : \Lambda \to [0, 1]$, each configuration receives a \emph{super-uniform $p$-value}, \emph{family-wise error control} covers the $m$ configurations tested at once, and the \emph{Learn-Then-Test} framework \citep{angelopoulos2021learn} combines them into a set of configurations whose risks are controlled simultaneously. Our $p$-value is an exact binomial tail, with no asymptotics and no distributional model. We detail the formalism and the notation in Appendix~\ref{app:formalism}.

\subsection{The serving-configuration space}
\label{sec:motivation-space}

A deployed block-diffusion engine has a multi-dimensional configuration, spanning accept and semi-completion thresholds, block-add and participation schedules, buffer depth, checkpoint and numeric precision. Published systems pick points in this space by hand. For example, LLaDA2.0 is released at one point with a reported $2.1\times$ speedup, LLaDA2.1 offers a menu of speed and quality modes, and SDAR reports a sweep over block sizes \citep{bie2025llada2,bie2026llada21,cheng2025sdar}.

Two observations about this space motivate the measurements that follow. First, on math decode traces of GSM8K and MATH-500, roughly $37\%$ of active block-forwards in both families are no-ops that commit no token, and most of them fall on freshly added slots that are still fully masked. Second, only about two blocks stay in flight. A four-block buffer has nearly four still-masked slots in its forward window on an average step, yet the scheduler admits only about two blocks, and at most two are partially resolved in over $99\%$ of steps in both families. Halving or doubling the buffer barely changes this saturation, which \citet{jin2026mbd} observe for naive multi-block decoding but leave unexplained. Appendix~\ref{app:engine} gives the measurements for each family and the buffer sweep.

\subsection{Three levers of the serving loop}
\label{sec:motivation-loci}

An intervention inside the serving loop can act at three points, and each point defines one lever. The training gradient can \emph{train} the model to commit more tokens in each forward \citep{chen2025dparallel,qian2026d3llm,wang2025d2f,lee2026answerconditioned}, the pre-forward state can \emph{allocate} forwards away from wasted slots \citep{lee2026dart_routing}, and the post-forward state decides which resolved tokens to \emph{commit} \citep{chen2026dmax,bie2026llada21}. Engine redesign and cross-request scheduling change the loop itself and lie outside our scope.

\section{Measuring the Three Levers}
\label{sec:cmech}

\begin{figure}[t]
\centering
\includegraphics[width=\linewidth]{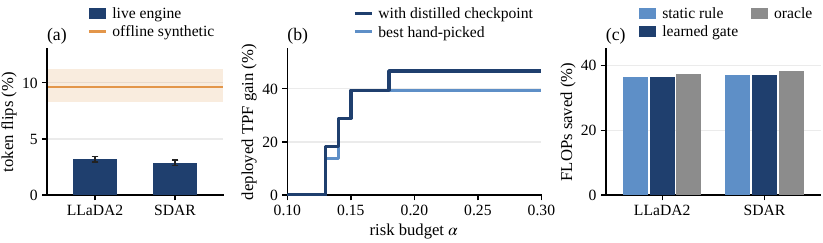}
\caption{Measurements of the serving loop. (a) Token flips when unresolved predecessors are finalized. (b) Deployed TPF gain of Redline on MATH-500 with the distilled checkpoint, against the best hand-picked configuration. (c) FLOPs saved in the compute-bound regime. Whiskers give $95\%$ intervals.}
\label{fig:probes}
\end{figure}

Figure~\ref{fig:probes} summarizes our measurements of the blocks in flight and of the three levers on the engine and its decode traces, and Appendix~\ref{app:prereg} gives the protocol and the control of each.

\subsection{Why only two blocks stay in flight}
\label{sec:cmech-depth}

Only about two blocks stay in flight because the masked slots of deeper blocks remain uncertain. We re-run each committing forward inside the engine's own decode traces with its unresolved predecessor finalized to its eventual tokens. As Figure~\ref{fig:probes}a shows, in both families this flips only about $3\%$ of the positions committed while an older block still holds masks, roughly a third of the offline rate on synthetically masked states. In the same SDAR traces, the masked slots of the deeper blocks carry more predictive entropy than those of the first active block in $96.7\%$ of the decode steps that hold a deeper block, and they still do in $67.3\%$ once every predecessor is finalized, which removes about three fifths of their excess entropy. During serving, only $2.6$ and $3.0\%$ of the masked slots in the second and third blocks reach the accept threshold of $0.95$, against $11.3\%$ in the first.

\subsection{The training lever: self-distillation on engine-decoded targets}
\label{sec:cmech-train}

\begingroup\setlength{\intextsep}{0pt}%
\begin{wraptable}{r}{0.5\textwidth}
\caption{Self-distillation on engine-decoded targets, SDAR on $256$ held-out GSM8K prompts. Engine accuracy and TPF at each admission threshold $\tau_{\text{add}}$, with the change in points and in percent.}
\label{tab:ctrain}
\centering\footnotesize\setlength{\tabcolsep}{2.5pt}
\begin{tabular*}{\linewidth}{@{\extracolsep{\fill}}l ccc ccc@{}}
\toprule
 & \multicolumn{3}{c}{\bfseries engine accuracy} & \multicolumn{3}{c}{\bfseries engine TPF} \\
\cmidrule(lr){2-4}\cmidrule(l){5-7}
\bfseries\boldmath $\tau_{\text{add}}$ & \bfseries 0.05 & \bfseries 0.10 & \bfseries 0.20 & \bfseries 0.05 & \bfseries 0.10 & \bfseries 0.20 \\
\midrule
base & 0.895 & 0.895 & 0.902 & 3.20 & 3.18 & 3.19 \\
distilled & 0.898 & 0.902 & 0.902 & 3.36 & 3.37 & 3.37 \\
change & $+0.39$ & $+0.78$ & $+0.00$ & $+5.2$ & $+5.8$ & $+5.8$ \\
\bottomrule
\end{tabular*}
\end{wraptable}

We distill the model toward committing more tokens in each forward, on multi-block targets that the engine decodes with its own confidence rule. Table~\ref{tab:ctrain} compares the distilled checkpoint with the base checkpoint on held-out GSM8K prompts. At every admission threshold, it gains $5.2$ to $5.8\%$ TPF while accuracy changes by $0.0$ to $+0.8$ points. We then serve it at the eight threshold settings of the SDAR math grid and test it beside the eight hand-picked configurations on the MATH-500 half of the math set, which neither the training nor the selection of the checkpoint touched. As Figure~\ref{fig:probes}b shows, at every risk budget from $0.18$ on, Redline deploys a configuration of the distilled checkpoint, with $+46.7\%$ TPF against $+39.5\%$ for the best hand-picked configuration on the same half.\par
\endgroup

\subsection{The allocation lever: a static skip rule}
\label{sec:cmech-alloc}

The no-op forwards of Section~\ref{sec:motivation-space} invite a participation schedule that skips them. In a replay of the engine's decode traces, a static rule that skips a fully masked block while an older block is still active saves $36$ to $37\%$ of FLOPs in the compute-bound regime. As Figure~\ref{fig:probes}c shows, this lies within $1.3$ points of an oracle that removes every zero-commit forward, and a learned participation gate matches the static rule to within $0.02$ points.

\subsection{The commitment lever: default commitment is maximal}
\label{sec:cmech-pbc}

Under the engine's default commit rule, every fully resolved block preceded only by committed blocks enters the cache in the same step, and no block that is safe to commit is left waiting. We prove this maximal-exact-commit invariant from the scheduler in Appendix~\ref{app:mec}, and it holds at all $281{,}005$ measured block-steps, whereas the revision-capable samplers of LLaDA2.1 and DMax \citep{bie2026llada21,chen2026dmax} can defer a completed block by one step.

\section{Risk-Controlled Configuration Selection}
\label{sec:ccert}

As Section~\ref{sec:cmech} shows, the default commit rule already commits every fully resolved block, and the masked slots of deeper blocks seldom reach the accept threshold. A configuration that is faster at a fixed checkpoint therefore accepts tokens at lower confidence, and it can fail on prompts that the reference answers correctly. We build Redline from the correctness of each answer on held-out calibration prompts, from which the risk of a fixed operating point can be bounded with finite-sample confidence and without distributional assumptions, provided that the calibration prompts are drawn i.i.d.\ from the deployment distribution. Because each outcome records the realized correctness of a served output, the risk also covers the engine's run-to-run variation.

\subsection{Risk and cost}
\label{sec:ccert-estimand}

For each candidate configuration $\lambda$ in the serving-configuration space, we define the reference-relative risk
\begin{equation}
R(\lambda) \;=\; \Pr_{x \sim \mathcal{D}}\bigl[\,\text{reference}(x)\ \text{correct} \,\wedge\, \lambda(x)\ \text{incorrect}\,\bigr],
\label{eq:risk}
\end{equation}
the probability that the reference answers a prompt correctly and $\lambda$ does not, which is the population form of the negative flip rate that model-update studies track \citep{yan2021positive,xie2021regression}. A user who is switched to $\lambda$ meets each such failure as a regression, whatever $\lambda$ fixes on other prompts \citep{bansal2019updates}. The risk also upper-bounds the net accuracy drop $R - G$, where $G = \Pr(\text{reference incorrect} \wedge \lambda\ \text{correct}) \ge 0$, and some faster configurations even gain net accuracy. Table~\ref{tab:unlock} reports the net accuracy change beside each gain.

The reference, the conservative baseline of each space, is the engine's default configuration, lossless verification or bf16 weights, depending on the method. The cost, whether TPF, accepted tokens for each target forward or weight memory, serves only to choose among the configurations that pass, so the guarantee attaches to $R$ alone, and we report the TPF \emph{of} the deployed configuration, not a guaranteed TPF. Where TPF and verbosity could mix, we also report the number of forwards for each request.

\subsection{Redline in four steps}
\label{sec:ccert-procedure}

Redline instantiates Learn-Then-Test \citep{angelopoulos2021learn} for serving, with the reference-relative joint risk of Eq.~\ref{eq:risk} in place of a marginal error rate. (i) \emph{Fix} a grid $\Lambda$ of $m$ configurations that includes the reference, $n$ i.i.d.\ calibration prompts, a risk budget $\alpha$ and a failure probability $\delta$, with $\delta = 0.10$ throughout. (ii) \emph{Score} each configuration by counting the prompts that the reference gets right and $\lambda$ gets wrong, $V_\lambda$, and form the one-sided exact binomial $p$-value $p_\lambda = \Pr[\mathrm{Bin}(n, \alpha) \le V_\lambda]$ for $H_0 : R(\lambda) > \alpha$. (iii) \emph{Test} with the Holm step-down procedure \citep{holm1979simple} at family-wise level $\delta$, which sorts $p_{(1)} \le \cdots \le p_{(m)}$ and rejects while $p_{(i)} \le \delta/(m - i + 1)$, stopping at the first failure. We call the configurations whose null hypothesis is rejected \emph{valid}. (iv) \emph{Deploy} the fastest valid configuration $\lambda^\star$, or for quantization the one with the least memory. Because Holm controls the family-wise error over all valid configurations at once, the deployed configuration satisfies
\begin{equation}
\Pr\bigl(R(\lambda^\star) \le \alpha\bigr) \;\ge\; 1 - \delta
\label{eq:guarantee}
\end{equation}
in finite samples and without distributional assumptions, whichever valid configuration the cost selects, as Appendix~\ref{app:deploy-simult} shows.

We use a Holm-class correction because shared calibration prompts make the $p$-values dependent in an uncontrolled way, and Holm tolerates arbitrary dependence. Measured risk is also non-monotone across the grid, since the configurations with the strictest accept threshold run slower than the reference yet still carry nonzero risk, so fixed-sequence testing, which walks a data-independent order, serves only as a secondary procedure. By construction, every configuration valid under Bonferroni is valid under Holm, and on our grids Holm deploys a faster configuration than Bonferroni at several budgets. Appendix~\ref{app:grids-frontier} compares the three corrections on every grid.

The guarantee holds at each budget with probability at least $1 - \delta$, and a similar bound covers the smallest budget at which a task gains speed, which we read off the grid of budgets. Gaining speed at a budget below the smallest true risk among the $F$ configurations faster than the reference requires one of them to clear a Holm threshold of at most $\delta/F$, which a union bound limits to probability $\delta$.

The calibration size sets how close to the budget a configuration can sit and still pass reliably. With eight configurations, one that sits two points under a budget of $\alpha = 0.05$ passes with probability at least $0.8$ from about $1{,}000$ prompts. Because nothing in Redline inspects diffusion internals, it needs only a reference configuration, a grid of lossy settings, correctness outcomes and a cost measure, so we also apply it unchanged to speculative-decoding acceptance and weight quantization.

\section{Experiments}
\label{sec:experiments}

\begin{table}[t]
\caption{Configurations that Redline deploys at three risk budgets ($\delta = 0.10$, Holm), with the gain over the reference in the group's cost measure and the net accuracy change in points. Below $\alpha_{\min}$, the smallest budget at which a cheaper configuration passes, the reference is deployed.}
\label{tab:unlock}
\centering\small\setlength{\tabcolsep}{2.5pt}
\begin{tabularx}{\textwidth}{@{}lccc*{6}{>{\raggedleft\arraybackslash}X}@{}}
\toprule
 & & & & \multicolumn{2}{c}{\boldmath$\alpha = 0.10$} & \multicolumn{2}{c}{\boldmath$\alpha = 0.15$} & \multicolumn{2}{c}{\boldmath$\alpha = 0.20$} \\
\cmidrule(lr){5-6}\cmidrule(lr){7-8}\cmidrule(l){9-10}
\textbf{grid} & \boldmath$n$ & \boldmath$m$ & \boldmath$\alpha_{\min}$ & \textbf{gain} & \boldmath$\Delta$\textbf{acc.} & \textbf{gain} & \boldmath$\Delta$\textbf{acc.} & \textbf{gain} & \boldmath$\Delta$\textbf{acc.} \\
\midrule
\grouprow{10}{Block-diffusion serving, TPF gain (\%)} \\
LLaDA2 math & $1{,}012$ & $7$ & $0.07$ & $+37.5$ & $-4.0$ & $+37.5$ & $-4.0$ & $+37.5$ & $-4.0$ \\
LLaDA2 math, large grid & $1{,}012$ & $50$ & $0.07$ & $+65.4$ & $-3.4$ & $+72.0$ & $-5.9$ & $+72.0$ & $-5.9$ \\
LLaDA2 code & $542$ & $8$ & $0.11$ & $0.0$ & $0.0$ & $+33.7$ & $-7.0$ & $+42.4$ & $-13.1$ \\
SDAR math & $1{,}012$ & $8$ & $0.10$ & $+13.1$ & $-2.5$ & $+36.6$ & $-4.6$ & $+36.6$ & $-4.6$ \\
SDAR code & $664$ & $8$ & $0.11$ & $0.0$ & $0.0$ & $+25.7$ & $-5.0$ & $+59.7$ & $-14.0$ \\
\grouprow{10}{Speculative decoding on GSM8K, gain in accepted tokens for each target forward (\%)} \\
Llama & $512$ & $7$ & $0.09$ & $+13.8$ & $-3.3$ & $+16.8$ & $-9.0$ & $+16.8$ & $-9.0$ \\
Qwen & $512$ & $7$ & $0.04$ & $+18.1$ & $-1.6$ & $+18.1$ & $-1.6$ & $+18.1$ & $-1.6$ \\
\grouprow{10}{Weight quantization on GSM8K, reduction in weight memory} \\
Llama & $512$ & $4$ & $0.07$ & $2.81\times$ & $-1.4$ & $2.81\times$ & $-1.4$ & $2.81\times$ & $-1.4$ \\
\bottomrule
\end{tabularx}
\end{table}

\subsection{Setup}
\label{sec:exp-setup}

\textbf{Models and engines.} We apply Redline to LLaDA2-mini \citep{bie2025llada2} and SDAR-8B \citep{cheng2025sdar} on the multi-block engine \citep{jin2026mbd}, with the SDAR accept threshold made configurable, each over an accept $\times$ semi-completion grid served under a $4{,}096$-token generation budget with greedy decoding. The reference of both families is the engine's default, accept $0.95$ with semi-completion $0.90$, and Appendix~\ref{app:engine-config} lists every configuration.

\textbf{Benchmarks and scoring.} Math is the same fixed composite in both families, GSM8K \citep{gsm8k} and MATH-500 \citep{math} with $n = 1{,}012$ prompts, and code pairs HumanEval+ with an MBPP variant, with $n = 542$ for LLaDA2 and $664$ for SDAR \citep{humaneval,mbpp,evalplus}. We score code by execution-based pass@1 and math by exact match.

\textbf{Autoregressive methods.} For speculative decoding, we test Llama-3.1-8B with a Llama-3.2-1B draft and Qwen2.5-7B with a Qwen2.5-1.5B draft on GSM8K ($n = 512$) under the typical acceptance of Medusa \citep{cai2024medusa}, against lossless verification, whose risk is zero by construction. For quantization, we test bitsandbytes int8, nf4 and fp4 weights \citep{dettmers2022int8,dettmers2023qlora} against bf16.

\subsection{Where each task gains speed}
\label{sec:exp-unlock}

\begin{figure}[t]
\centering
\includegraphics[width=\linewidth]{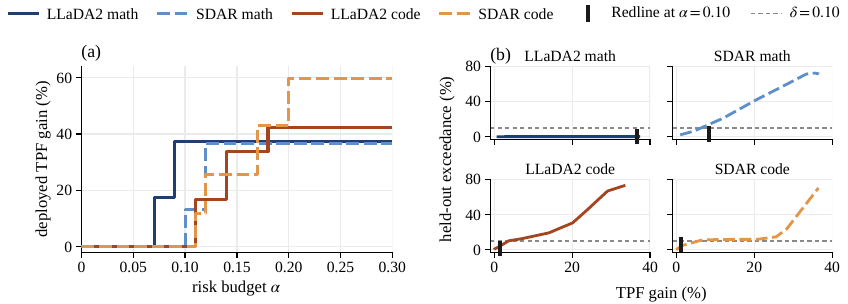}
\caption{Redline and the mean-accuracy rule. (a) Deployed TPF gain across risk budgets. (b) Held-out exceedance, the rate at which the deployed configuration exceeds the budget on held-out halves, against TPF gain at $\alpha = 0.10$, for the mean-accuracy rule across its tolerances and for Redline.}
\label{fig:alpha-curve}
\end{figure}

Table~\ref{tab:unlock} shows the deployed configurations at three budgets, and Figure~\ref{fig:alpha-curve}a traces the deployed gain over the whole range of budgets. In both families, math gains speed at a smaller budget than code, and at every tested budget at which code gains speed, math does too. Every nonzero gain in TPF or accepted tokens carries a $95\%$ paired-bootstrap interval that excludes zero. Both math configurations deployed at $\alpha = 0.10$ stay the same when every grid is tested again at $\delta = 0.05$ or $0.01$, and hence also under a Bonferroni split of $\delta$ over the four grids.

At $\alpha = 0.10$, the deployed math configurations already run well ahead of the reference, at $+37.5\%$ TPF on LLaDA2 and $+13.1\%$ on SDAR, with $95\%$ intervals of $34.0$ to $41.1\%$ and $7.1$ to $19.6\%$. Both sit about three points of empirical risk under the budget, and the SDAR math configuration also cuts the forwards of each request by $15.9\%$.

Code first gains speed at $\alpha = 0.11$ in both families, and the ordering holds configuration by configuration. Each of the five configurations faster than the reference fails on $1.7$ to $2.9$ times as large a share of the reference's correct answers on code as on math. With math subsampled to the size of the code set, the smallest math budget with a gain stays below the code one in every LLaDA2 draw and at or below it in $99.1\%$ of SDAR draws. Appendices~\ref{app:grids-tables}, \ref{app:grids-frontier} and~\ref{app:grids-nfe} give the intervals, both comparisons and the forwards.

\subsection{Against the mean-accuracy rule}
\label{sec:exp-breadth}

Figure~\ref{fig:alpha-curve}b compares Redline with the rule it replaces, selection by mean accuracy, on the same grids and prompts. A mean lets the prompts that a configuration fixes offset those it fails, and the SDAR math configuration deployed at $\alpha = 0.10$, for instance, turns a correct reference answer into a wrong one on $7.1\%$ of prompts, nearly three times its net accuracy drop. Over $1{,}000$ random half splits of each calibration set, we apply each rule to one half and score its deployed configuration on the other. We call the rate at which the risk $R$ of that configuration exceeds the budget, on the test half or pooled over all prompts, the exceedance.

The mean-accuracy rule deploys the fastest configuration whose calibration-half accuracy is within $t$ points of the reference's, for $t \in \{0, 1, \ldots, 8, 10\}$. At $\alpha = 0.10$, no single tolerance is both as fast as Redline on LLaDA2 math and as rarely over the budget on SDAR math. Every tolerance up to four points gains less TPF than Redline on LLaDA2 math, and every tolerance from five points up exceeds the budget in over $70\%$ of held-out splits on SDAR math, against $3.4\%$ for Redline. The trade-off persists when each rule is applied to $30$ or $70\%$ of the prompts instead of half. A tolerance chosen separately for each grid could be checked against the budget only by measuring $R$ on that grid, which a mean does not report.

Each ingredient of Redline is needed. Dropping the multiplicity correction raises the held-out exceedance on SDAR math at $\alpha = 0.10$ from $3.4$ to $17.5\%$, and dropping the finite-sample margin as well, the plug-in rule $\hat R \le \alpha$, exceeds the budget in up to $64.1\%$ of held-out splits. Testing the net drop instead of the joint risk, with the same Holm step and a betting $p$-value \citep{waudbysmith2024betting}, keeps that drop within budget in every split on SDAR math at $\alpha = 0.10$, yet its deployed configurations exceed the joint budget in $91.4\%$ of splits against the pooled risk. Redline itself stays under $7\%$ held-out exceedance and at or under $0.2\%$ pooled for all sixteen pairs of grid and budget. Appendix~\ref{app:mean-selector} lists every pair and every rule.

\subsection{Speculative decoding and quantization}
\label{sec:exp-arknobs}

Table~\ref{tab:unlock} also lists the deployed configurations of two public autoregressive methods. For the typical acceptance of Medusa, a faster setting is deployed from $\alpha = 0.04$ on for the Qwen pair, with a net accuracy gain at $\alpha = 0.05$, and from $\alpha = 0.09$ on for the Llama pair. The Llama draft agrees with the target's argmax on more tokens than the Qwen draft, $90.5\%$ against $88.7\%$, yet needs a larger budget, since the risk measures whether a divergent token changes the final answer rather than how often tokens diverge. Under quantization, all three low-precision formats are valid at $\alpha = 0.10$, and the deployed format, nf4, reduces weight memory by a measured factor of $2.81$ and peak memory during generation by $2.75$. Appendix~\ref{app:generality} details both methods.

\subsection{Validity and larger grids}
\label{sec:exp-validity}

\begin{figure}[t]
\centering
\includegraphics[width=\linewidth]{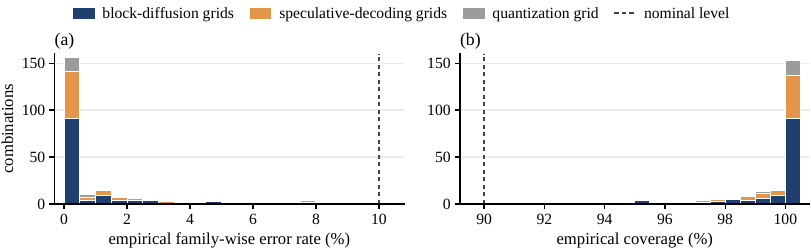}
\caption{Out-of-sample validity of Redline over $220$ combinations of grid, risk budget and calibration size, each on $1{,}000$ random splits. (a) Empirical family-wise error rate. (b) Empirical coverage.}
\label{fig:coverage}
\end{figure}

\begin{figure}[t]
\centering
\includegraphics[width=\linewidth]{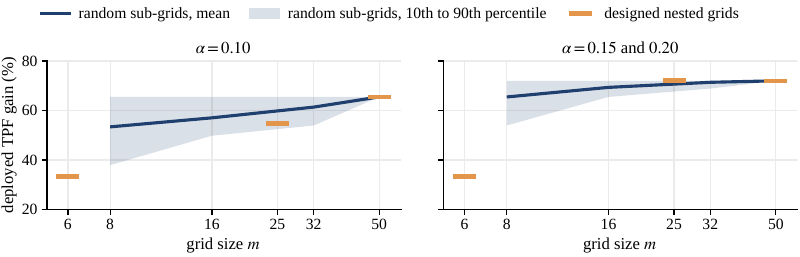}
\caption{Deployed TPF gain of Redline against grid size on the LLaDA2 math grid of $50$ configurations ($n = 1{,}012$, Holm, $\delta = 0.10$), for $1{,}000$ random sub-grids of each size that hold the reference and for the designed nested grids.}
\label{fig:gridsize}
\end{figure}

As Figure~\ref{fig:coverage} shows, across $12$ grids and $220$ combinations, repeated random calibration and test splits give a minimum coverage of $0.904$ and a maximum family-wise error rate of $0.096$, and no combination exceeds the nominal level. When we run Redline on one random half of each calibration set, the deployed configurations re-measured on the other half stay close to their in-sample gains, for instance $+37.6\%$ against $+37.5\%$ TPF for the LLaDA2 math configuration at $\alpha = 0.10$ and $+13.4\%$ against $+13.1\%$ for the SDAR math configuration, with net accuracy changes within a point of the in-sample values. Appendices~\ref{app:validity} and~\ref{app:heldout-deploy} give the split protocol.

A joint grid over the accept, semi-completion and add thresholds deploys exactly the LLaDA2 math configuration of Table~\ref{tab:unlock}, and on Fast-dLLM-v2 a faster accept threshold is deployed at $+15.3\%$ TPF at $\alpha = 0.10$. On a separate LLaDA2 math grid of $50$ configurations over all three lossy thresholds of the engine, listed in Table~\ref{tab:unlock} with its own reference, $36$ of the $50$ configurations are valid at $\alpha = 0.10$ and all $50$ at $\alpha \ge 0.15$. Its deployed configurations reach $+65.4\%$ and $+72.0\%$ TPF, with a held-out exceedance of $5.3\%$ at $\alpha = 0.10$ and none at $0.15$ and $0.20$, where every split deploys the same configuration. As Figure~\ref{fig:gridsize} shows, random sub-grids deploy faster configurations as they grow, despite the stricter correction. Appendices~\ref{app:grids-3d}, \ref{app:grids-absorption} and~\ref{app:grids-biggrid} give these three grids.

\section{Related Work}
\label{sec:related}

\textbf{Block-diffusion LMs.} Masked-diffusion LMs \citep{sahoo2024mdlm,nie2025llada,ye2025dream} established the discrete-diffusion alternative to autoregressive decoding, and BD3-LM \citep{arriola2025block} is the architectural ancestor of our setting. SDAR \citep{cheng2025sdar} and LLaDA2.0 and LLaDA2.1 \citep{bie2025llada2,bie2026llada21} are the families we evaluate, and their published operating points come without a prompt-level guarantee. We serve on the multi-block engine of \citet{jin2026mbd}.

\textbf{Efficient and parallel decoding for diffusion LMs.} Fast-dLLM and Fast-dLLM-v2 \citep{wu2025fastdllm,wu2025fastdllmv2}, D2F \citep{wang2025d2f}, AdaBlock-dLLM \citep{lu2025adablock}, dParallel \citep{chen2025dparallel}, LoPA \citep{xu2025lopa}, d3LLM \citep{qian2026d3llm}, LightningRL \citep{hu2026lightningrl} and DMax \citep{chen2026dmax} push the same frontier by training, scheduling or caching. None of them states a distribution-free finite-sample bound on deployed quality or controls multiplicity over a configuration space. The nearest statement is Theorem~1 of Fast-dLLM, which is deterministic, conditional on confidence and has no $\delta$. The thresholds and checkpoints of these methods are \emph{inputs} to our space rather than alternatives to it.

\textbf{Distribution-free risk control.} Learn-Then-Test \citep{angelopoulos2021learn} is our testing framework, and conformal risk control \citep{angelopoulos2022conformal} bounds the expected loss rather than the probability of exceeding the budget. CALM \citep{schuster2022confident}, the closest prior use in decoding, calibrates one early-exit threshold of autoregressive decoding against the full model by a fixed-sequence walk. Redline selects from a searched grid of serving configurations whose risk is non-monotone in the thresholds, which Holm handles without an order, and applies unchanged to any cost measure. To our knowledge, no concurrent work selects serving configurations by distribution-free multiple testing, and the nearest analogs are \citet{farzaneh2026statistically} and \citet{cai2026confidence}.

\textbf{Regressions under model updates.} Model-update studies train a new model to lower its negative flip rate against an old one \citep{yan2021positive,xie2021regression}. We instead bound that rate below a budget for the serving configurations of a fixed model, without training.

\textbf{Lossy autoregressive acceleration.} Speculative decoding \citep{leviathan2023fast,chen2023accelerating} and the typical acceptance of Medusa \citep{cai2024medusa} supply the acceptance setting and quantization the memory setting, both of which Redline calibrates. Appendix~\ref{app:relatedmatrix} compares these methods with our guarantee.

\section{Discussion \& Conclusion}
\label{sec:discussion}

\textbf{Limitations.}\label{sec:limitations} The guarantee is relative to the calibration distribution and to a reference configuration, so a deployment that serves a different prompt distribution or batching policy re-calibrates first. Our experiments cover greedy decoding on math and code in two block-diffusion families, and Redline applies to any task with verifiable correctness. Each guarantee holds at its own level $1 - \delta$, and several guarantees hold jointly when $\delta$ is split across them.

\textbf{Conclusion.} Redline makes the operating points of an engine, hand-picked or learned, selectable at a stated risk. A checkpoint trained with any signal, such as self-distillation or verifier-rewarded reinforcement learning, joins the grid as one more set of configurations, and Redline itself needs no training and carries over unchanged to any lossy serving method with a cost measure. On four grids it orders math before code in the same way across two families, and it bounds the regressions that mean accuracy hides behind fixes.

\section*{Ethics Statement}

We select serving configurations of existing open-weight language models and introduce no new model, dataset or study with human subjects. All benchmarks are public (GSM8K, MATH-500, HumanEval, HumanEval+, MBPP and MBPP+) and are used under their licenses. The guarantee holds for the calibrated prompt distribution, and a deployment that serves a different distribution re-calibrates before relying on it.

\bibliography{references}

@inproceedings{arriola2025block,
  title     = {Block Diffusion: Interpolating Between Autoregressive and Diffusion Language Models},
  author    = {Marianne Arriola and Aaron Gokaslan and Justin T. Chiu and Zhihan Yang and Zhixuan Qi and Jiaqi Han and Subham Sekhar Sahoo and Volodymyr Kuleshov},
  booktitle = {International Conference on Learning Representations (ICLR)},
  year      = {2025},
  url       = {https://proceedings.iclr.cc/paper_files/paper/2025/hash/7ede97c3e082c6df10a8d6103a2eebd2-Abstract-Conference.html},
  eprint    = {2503.09573},
  archivePrefix = {arXiv}
}

@inproceedings{cheng2025sdar,
  title     = {{SDAR}: A Synergistic Diffusion-AutoRegression Paradigm for Scalable Sequence Generation},
  author    = {Shuang Cheng and Yihan Bian and Dawei Liu and Yuhua Jiang and Yihao Liu and Linfeng Zhang and Qian Yao and Zhongbo Tian and Wenhai Wang and Qipeng Guo and Kai Chen and Biqing Qi and Bowen Zhou},
  booktitle = {Findings of the Association for Computational Linguistics: ACL 2026},
  pages     = {22058--22075},
  year      = {2026},
  doi       = {10.18653/v1/2026.findings-acl.1110},
  eprint    = {2510.06303},
  archivePrefix = {arXiv}
}

@article{bie2025llada2,
  title   = {{LLaDA2.0}: Scaling Up Diffusion Language Models to {100B}},
  author  = {Tiwei Bie and Maosong Cao and Kun Chen and Lun Du and Mingliang Gong and Zhuochen Gong and Yanmei Gu and Jiaqi Hu and Zenan Huang and Zhenzhong Lan and Chengxi Li and others},
  journal = {arXiv preprint arXiv:2512.15745},
  year    = {2025},
  url     = {https://arxiv.org/abs/2512.15745}
}

@article{bie2026llada21,
  title   = {{LLaDA2.1}: Speeding Up Text Diffusion via Token Editing},
  author  = {Tiwei Bie and Maosong Cao and Xiang Cao and Bingsen Chen and Fuyuan Chen and Kun Chen and Lun Du and Daozhuo Feng and Haibo Feng and Mingliang Gong and Zhuocheng Gong and others},
  journal = {arXiv preprint arXiv:2602.08676},
  year    = {2026},
  url     = {https://arxiv.org/abs/2602.08676}
}

@article{jin2026mbd,
  title   = {Multi-Block Diffusion Language Models},
  author  = {Yijie Jin and Jiajun Xu and Yuxuan Liu and Chenkai Xu and Yi Tu and Jiajun Li and Dandan Tu and Xiaohui Yan and Kai Yu and Pengfei Liu and Zhijie Deng},
  journal = {arXiv preprint arXiv:2606.29215},
  year    = {2026},
  url     = {https://arxiv.org/abs/2606.29215}
}

@inproceedings{nie2025llada,
  title     = {Large Language Diffusion Models},
  author    = {Shen Nie and Fengqi Zhu and Zebin You and Xiaolu Zhang and Jingyang Ou and Jun Hu and Jun Zhou and Yankai Lin and Ji-Rong Wen and Chongxuan Li},
  booktitle = {Advances in Neural Information Processing Systems (NeurIPS)},
  year      = {2025},
  url       = {https://proceedings.neurips.cc/paper_files/paper/2025/hash/48b383b24230e0e6e649d9c98dae4d8c-Abstract-Conference.html},
  eprint    = {2502.09992},
  archivePrefix = {arXiv}
}

@article{ye2025dream,
  title   = {Dream {7B}: Diffusion Large Language Models},
  author  = {Jiacheng Ye and Zhihui Xie and Lin Zheng and Jiahui Gao and Zirui Wu and Xin Jiang and Zhenguo Li and Lingpeng Kong},
  journal = {arXiv preprint arXiv:2508.15487},
  year    = {2025},
  url     = {https://arxiv.org/abs/2508.15487}
}

@inproceedings{sahoo2024mdlm,
  title     = {Simple and Effective Masked Diffusion Language Models},
  author    = {Subham Sekhar Sahoo and Marianne Arriola and Yair Schiff and Aaron Gokaslan and Edgar Marroquin and Justin T. Chiu and Alexander Rush and Volodymyr Kuleshov},
  booktitle = {Advances in Neural Information Processing Systems (NeurIPS)},
  year      = {2024},
  url       = {https://proceedings.neurips.cc/paper_files/paper/2024/hash/eb0b13cc515724ab8015bc978fdde0ad-Abstract-Conference.html},
  eprint    = {2406.07524},
  archivePrefix = {arXiv}
}

@inproceedings{wu2025fastdllm,
  title     = {{Fast-dLLM}: Training-free Acceleration of Diffusion {LLM} by Enabling {KV} Cache and Parallel Decoding},
  author    = {Chengyue Wu and Hao Zhang and Shuchen Xue and Zhijian Liu and Shizhe Diao and Ligeng Zhu and Ping Luo and Song Han and Enze Xie},
  booktitle = {International Conference on Learning Representations (ICLR)},
  year      = {2026},
  url       = {https://proceedings.iclr.cc/paper_files/paper/2026/hash/5d8d4e6061c3ba96c240b7fa1ae3471d-Abstract-Conference.html},
  eprint    = {2505.22618},
  archivePrefix = {arXiv}
}

@inproceedings{wu2025fastdllmv2,
  title     = {{Fast-dLLM v2}: Efficient Block-Diffusion {LLM}},
  author    = {Chengyue Wu and Hao Zhang and Shuchen Xue and Shizhe Diao and Yonggan Fu and Zhijian Liu and Pavlo Molchanov and Ping Luo and Song Han and Enze Xie},
  booktitle = {International Conference on Learning Representations (ICLR)},
  year      = {2026},
  url       = {https://proceedings.iclr.cc/paper_files/paper/2026/hash/d0865cbe51d35ec322f9af9db7806fc7-Abstract-Conference.html},
  eprint    = {2509.26328},
  archivePrefix = {arXiv}
}

@inproceedings{wang2025d2f,
  title     = {Diffusion {LLM}s Can Do Faster-Than-{AR} Inference via Discrete Diffusion Forcing},
  author    = {Xu Wang and Chenkai Xu and Yijie Jin and Jiachun Jin and Hao Zhang and Kai Yu and Zhijie Deng},
  booktitle = {International Conference on Learning Representations (ICLR)},
  year      = {2026},
  url       = {https://proceedings.iclr.cc/paper_files/paper/2026/hash/04bb76a98d9f32226b3beba7bd26a51f-Abstract-Conference.html},
  eprint    = {2508.09192},
  archivePrefix = {arXiv}
}

@inproceedings{lu2025adablock,
  title     = {{AdaBlock-dLLM}: Semantic-Aware Diffusion {LLM} Inference via Adaptive Block Size},
  author    = {Guanxi Lu and Hao Mark Chen and Yuto Karashima and Zhican Wang and Daichi Fujiki and Hongxiang Fan},
  booktitle = {International Conference on Learning Representations (ICLR)},
  year      = {2026},
  url       = {https://proceedings.iclr.cc/paper_files/paper/2026/hash/e2eeb89acc98e8e506d719e330cbc43a-Abstract-Conference.html},
  eprint    = {2509.26432},
  archivePrefix = {arXiv}
}

@inproceedings{chen2025dparallel,
  title     = {{dParallel}: Learnable Parallel Decoding for {dLLM}s},
  author    = {Zigeng Chen and Gongfan Fang and Xinyin Ma and Ruonan Yu and Xinchao Wang},
  booktitle = {International Conference on Learning Representations (ICLR)},
  year      = {2026},
  url       = {https://proceedings.iclr.cc/paper_files/paper/2026/hash/57250222014c35949476f3f272c322d2-Abstract-Conference.html},
  eprint    = {2509.26488},
  archivePrefix = {arXiv}
}

@article{xu2025lopa,
  title   = {{LoPA}: Scaling {dLLM} Inference via Lookahead Parallel Decoding},
  author  = {Chenkai Xu and Yijie Jin and Jiajun Li and Yi Tu and Guoping Long and Dandan Tu and Mingcong Song and Hongjie Si and Tianqi Hou and Junchi Yan and Zhijie Deng},
  journal = {arXiv preprint arXiv:2512.16229},
  year    = {2025},
  url     = {https://arxiv.org/abs/2512.16229}
}

@inproceedings{qian2026d3llm,
  title     = {{d3LLM}: Ultra-Fast Diffusion {LLM} using Pseudo-Trajectory Distillation},
  author    = {Yu-Yang Qian and Junda Su and Lanxiang Hu and Peiyuan Zhang and Zhijie Deng and Peng Zhao and Hao Zhang},
  booktitle = {International Conference on Machine Learning (ICML)},
  year      = {2026},
  url       = {https://icml.cc/virtual/2026/poster/61269},
  eprint    = {2601.07568},
  archivePrefix = {arXiv}
}

@inproceedings{hu2026lightningrl,
  title     = {{LightningRL}: Breaking the Accuracy--Parallelism Trade-off of Block-wise {dLLM}s via Reinforcement Learning},
  author    = {Yanzhe Hu and Yijie Jin and Pengfei Liu and Kai Yu and Zhijie Deng},
  booktitle = {International Conference on Machine Learning (ICML)},
  year      = {2026},
  url       = {https://icml.cc/virtual/2026/poster/65221},
  eprint    = {2603.13319},
  archivePrefix = {arXiv}
}

@article{chen2026dmax,
  title   = {{DMax}: Aggressive Parallel Decoding for {dLLM}s},
  author  = {Zigeng Chen and Gongfan Fang and Xinyin Ma and Ruonan Yu and Xinchao Wang},
  journal = {arXiv preprint arXiv:2604.08302},
  year    = {2026},
  url     = {https://arxiv.org/abs/2604.08302}
}

@article{angelopoulos2021learn,
  title   = {Learn then Test: Calibrating Predictive Algorithms to Achieve Risk Control},
  author  = {Anastasios N. Angelopoulos and Stephen Bates and Emmanuel J. Cand{\`e}s and Michael I. Jordan and Lihua Lei},
  journal = {The Annals of Applied Statistics},
  volume  = {19},
  number  = {2},
  pages   = {1641--1662},
  year    = {2025},
  doi     = {10.1214/24-AOAS1998},
  eprint  = {2110.01052},
  archivePrefix = {arXiv}
}

@article{holm1979simple,
  title   = {A Simple Sequentially Rejective Multiple Test Procedure},
  author  = {Sture Holm},
  journal = {Scandinavian Journal of Statistics},
  volume  = {6},
  number  = {2},
  pages   = {65--70},
  year    = {1979}
}

@inproceedings{yan2021positive,
  title     = {Positive-Congruent Training: Towards Regression-Free Model Updates},
  author    = {Sijie Yan and Yuanjun Xiong and Kaustav Kundu and Shuo Yang and Siqi Deng and Meng Wang and Wei Xia and Stefano Soatto},
  booktitle = {Proceedings of the IEEE/CVF Conference on Computer Vision and Pattern Recognition (CVPR)},
  pages     = {14294--14303},
  year      = {2021},
  doi       = {10.1109/CVPR46437.2021.01407}
}

@inproceedings{xie2021regression,
  title     = {Regression Bugs Are In Your Model! {M}easuring, Reducing and Analyzing Regressions In {NLP} Model Updates},
  author    = {Yuqing Xie and Yi-An Lai and Yuanjun Xiong and Yi Zhang and Stefano Soatto},
  booktitle = {Proceedings of the 59th Annual Meeting of the Association for Computational Linguistics and the 11th International Joint Conference on Natural Language Processing (ACL-IJCNLP)},
  pages     = {6589--6602},
  year      = {2021},
  doi       = {10.18653/v1/2021.acl-long.515}
}

@inproceedings{bansal2019updates,
  title     = {Updates in Human-{AI} Teams: Understanding and Addressing the Performance/Compatibility Tradeoff},
  author    = {Gagan Bansal and Besmira Nushi and Ece Kamar and Daniel S. Weld and Walter S. Lasecki and Eric Horvitz},
  booktitle = {Proceedings of the AAAI Conference on Artificial Intelligence (AAAI)},
  volume    = {33},
  pages     = {2429--2437},
  year      = {2019},
  doi       = {10.1609/aaai.v33i01.33012429}
}

@inproceedings{schuster2022confident,
  title     = {Confident Adaptive Language Modeling},
  author    = {Tal Schuster and Adam Fisch and Jai Gupta and Mostafa Dehghani and Dara Bahri and Vinh Q. Tran and Yi Tay and Donald Metzler},
  booktitle = {Advances in Neural Information Processing Systems (NeurIPS)},
  year      = {2022},
  url       = {https://arxiv.org/abs/2207.07061}
}

@inproceedings{angelopoulos2022conformal,
  title     = {Conformal Risk Control},
  author    = {Anastasios N. Angelopoulos and Stephen Bates and Adam Fisch and Lihua Lei and Tal Schuster},
  booktitle = {International Conference on Learning Representations (ICLR)},
  year      = {2024},
  url       = {https://proceedings.iclr.cc/paper_files/paper/2024/hash/f3549ef9b5ff520a7e41ff3cc306ab2b-Abstract-Conference.html},
  eprint    = {2208.02814},
  archivePrefix = {arXiv}
}

@article{farzaneh2026statistically,
  title   = {Statistically Valid Post-Training Hyperparameter Selection: From Tuning to Guarantees},
  author  = {Amirmohammad Farzaneh and Osvaldo Simeone},
  journal = {arXiv preprint arXiv:2606.25601},
  year    = {2026},
  url     = {https://arxiv.org/abs/2606.25601}
}

@article{cai2026confidence,
  title   = {Confidence-Based Decoding is Provably Efficient for Diffusion Language Models},
  author  = {Changxiao Cai and Gen Li},
  journal = {arXiv preprint arXiv:2603.22248},
  year    = {2026},
  url     = {https://arxiv.org/abs/2603.22248}
}

@inproceedings{leviathan2023fast,
  title     = {Fast Inference from Transformers via Speculative Decoding},
  author    = {Yaniv Leviathan and Matan Kalman and Yossi Matias},
  booktitle = {International Conference on Machine Learning (ICML)},
  year      = {2023},
  url       = {https://proceedings.mlr.press/v202/leviathan23a.html},
  eprint    = {2211.17192},
  archivePrefix = {arXiv}
}

@article{chen2023accelerating,
  title   = {Accelerating Large Language Model Decoding with Speculative Sampling},
  author  = {Charlie Chen and Sebastian Borgeaud and Geoffrey Irving and Jean-Baptiste Lespiau and Laurent Sifre and John Jumper},
  journal = {arXiv preprint arXiv:2302.01318},
  year    = {2023},
  url     = {https://arxiv.org/abs/2302.01318}
}

@inproceedings{cai2024medusa,
  title     = {Medusa: Simple {LLM} Inference Acceleration Framework with Multiple Decoding Heads},
  author    = {Tianle Cai and Yuhong Li and Zhengyang Geng and Hongwu Peng and Jason D. Lee and Deming Chen and Tri Dao},
  booktitle = {International Conference on Machine Learning (ICML)},
  year      = {2024},
  url       = {https://proceedings.mlr.press/v235/cai24b.html},
  eprint    = {2401.10774},
  archivePrefix = {arXiv}
}

@inproceedings{dettmers2022int8,
  title     = {{LLM.int8()}: 8-bit Matrix Multiplication for Transformers at Scale},
  author    = {Tim Dettmers and Mike Lewis and Younes Belkada and Luke Zettlemoyer},
  booktitle = {Advances in Neural Information Processing Systems (NeurIPS)},
  year      = {2022},
  url       = {https://arxiv.org/abs/2208.07339}
}

@inproceedings{dettmers2023qlora,
  title     = {{QLoRA}: Efficient Finetuning of Quantized {LLM}s},
  author    = {Tim Dettmers and Artidoro Pagnoni and Ari Holtzman and Luke Zettlemoyer},
  booktitle = {Advances in Neural Information Processing Systems (NeurIPS)},
  year      = {2023},
  url       = {https://arxiv.org/abs/2305.14314}
}

@article{lee2026acoer,
  title   = {Beyond Penalizing Mistakes: Stabilizing Efficiency Training in Large Reasoning Models via Adaptive Correct-Only Rewards},
  author  = {Jungseob Lee and Seungyoon Lee and Seongtae Hong and Minhyuk Kim and Chanjun Park and Heuiseok Lim},
  journal = {arXiv preprint arXiv:2606.22716},
  year    = {2026},
  url     = {https://arxiv.org/abs/2606.22716}
}

@article{lee2026dart_routing,
  title   = {{DART}: Draft-Agreement Routing for Training-Free Adaptive Thinking Budgets in Hybrid Reasoning Models},
  author  = {Jungseob Lee and Seongtae Hong and Seungjun Lee and Jaehyung Seo and Junyoung Son and Sugyeong Eo and Chanjun Park and Hyeongju Park and Hyeonseok Moon and Heuiseok Lim},
  journal = {arXiv preprint arXiv:2606.23181},
  year    = {2026},
  url     = {https://arxiv.org/abs/2606.23181}
}

@article{lee2026answerconditioned,
  title   = {Answer-Conditioned Chains of Thought Degrade Verifiable-Reasoning Distillation in Large Language Models},
  author  = {Jungseob Lee and Seungyoon Lee and Suhyune Son and Dongyub Jude Lee and Sungbin Han and Sugyeong Eo and Heuiseok Lim},
  journal = {arXiv preprint arXiv:2607.14552},
  year    = {2026},
  url     = {https://arxiv.org/abs/2607.14552}
}

@article{gsm8k,
  title   = {Training Verifiers to Solve Math Word Problems},
  author  = {Cobbe, Karl and Kosaraju, Vineet and Bavarian, Mohammad and Chen, Mark and Jun, Heewoo and Kaiser, Lukasz and Plappert, Matthias and Tworek, Jerry and Hilton, Jacob and Nakano, Reiichiro and Hesse, Christopher and Schulman, John},
  journal = {arXiv preprint arXiv:2110.14168},
  year    = {2021}
}

@inproceedings{math,
  title     = {Measuring Mathematical Problem Solving with the {MATH} Dataset},
  author    = {Hendrycks, Dan and Burns, Collin and Kadavath, Saurav and Arora, Akul and Basart, Steven and Tang, Eric and Song, Dawn and Steinhardt, Jacob},
  booktitle = {Proceedings of the Neural Information Processing Systems Track on Datasets and Benchmarks},
  year      = {2021},
  url       = {https://datasets-benchmarks-proceedings.neurips.cc/paper/2021/hash/be83ab3ecd0db773eb2dc1b0a17836a1-Abstract-round2.html},
  eprint    = {2103.03874},
  archivePrefix = {arXiv}
}

@article{humaneval,
  title   = {Evaluating Large Language Models Trained on Code},
  author  = {Chen, Mark and Tworek, Jerry and Jun, Heewoo and Yuan, Qiming and Pinto, Henrique Ponde de Oliveira and Kaplan, Jared and Edwards, Harri and Burda, Yuri and Joseph, Nicholas and Brockman, Greg and others},
  journal = {arXiv preprint arXiv:2107.03374},
  year    = {2021}
}

@article{mbpp,
  title   = {Program Synthesis with Large Language Models},
  author  = {Austin, Jacob and Odena, Augustus and Nye, Maxwell and Bosma, Maarten and Michalewski, Henryk and Dohan, David and Jiang, Ellen and Cai, Carrie and Terry, Michael and Le, Quoc and others},
  journal = {arXiv preprint arXiv:2108.07732},
  year    = {2021}
}

@article{evalplus,
  title   = {Is Your Code Generated by {ChatGPT} Really Correct? Rigorous Evaluation of Large Language Models for Code Generation},
  author  = {Liu, Jiawei and Xia, Chunqiu Steven and Wang, Yuyao and Zhang, Lingming},
  journal = {Advances in Neural Information Processing Systems},
  volume  = {36},
  pages   = {21558--21572},
  year    = {2023}
}

@article{waudbysmith2024betting,
  author  = {Ian Waudby-Smith and Aaditya Ramdas},
  title   = {Estimating means of bounded random variables by betting},
  journal = {Journal of the Royal Statistical Society Series B: Statistical Methodology},
  volume  = {86},
  number  = {1},
  pages   = {1--27},
  year    = {2024},
  doi     = {10.1093/jrsssb/qkad009}
}

@article{bates2021rcps,
  author  = {Stephen Bates and Anastasios Angelopoulos and Lihua Lei and Jitendra Malik and Michael I. Jordan},
  title   = {Distribution-free, risk-controlling prediction sets},
  journal = {Journal of the ACM},
  volume  = {68},
  number  = {6},
  year    = {2021},
  doi     = {10.1145/3478535}
}
\bibliographystyle{iclr2027_conference}

\clearpage
\appendix
\section{Risk-Control Formalism and Validity}
\label{app:formalism}

\begin{table}[t]
\caption{Notation used throughout the paper.}
\label{tab:notation}
\begin{center}\small
\begin{tabular}{@{}p{0.169\linewidth}p{0.80\linewidth}@{}}
\toprule
$\lambda$, $\Lambda$ & a serving configuration and the finite grid of configurations searched, spanning accept and semi-completion thresholds, block-add schedule, buffer depth, checkpoint and precision \\
reference & the conservative baseline configuration of each grid, which is the engine's default for block diffusion, lossless verification for speculative decoding and bf16 weights for quantization \\
$R(\lambda)$ & reference-relative risk, the \emph{joint} probability that the reference answers a prompt correctly and $\lambda$ does not. It upper-bounds the net accuracy drop \\
$\alpha$ & the risk budget. The guarantee asserts $R(\lambda) \le \alpha$ \\
$\delta$ & the failure probability. All statements about one grid hold jointly with probability at least $1 - \delta$, and $\delta = 0.10$ throughout \\
TPF & tokens committed in each model forward, a count ratio independent of host speed. The speculative-decoding analogue is the number of accepted tokens for each target forward \\
$n$ & the number of calibration prompts of a grid \\
acc85/semi70 & a configuration label, here accept threshold $0.85$ and semi-completion threshold $0.70$. A suffix such as add0.30 gives the admission threshold \\
\bottomrule
\end{tabular}
\end{center}
\end{table}

\subsection{Setup and the guarantee}
\label{app:setup}

Let $\Lambda = \{\lambda_1, \ldots, \lambda_m\}$ be the finite grid of serving configurations, evaluated on $n$ i.i.d.\ calibration prompts. Each configuration carries the bounded prompt-level loss of Section~\ref{sec:ccert-estimand}, the $0/1$ reference-relative joint loss with $L_i(\lambda) = 1$ exactly when the reference decode is correct on prompt $i$ and the decode of $\lambda$ is not, and the risk is $R(\lambda) = \mathbb{E}[L(\lambda)]$. The loss is the realized correctness of the served output, so the expectation runs over the prompt and over the serving randomness of the engine under the serving policy of the calibration runs, batch composition included. The i.i.d.\ assumption is on prompts drawn together with their realized outcomes. The guarantee therefore covers the configuration as it is served under that policy, and a different batching policy is a distribution shift that is re-calibrated like any other. Redline follows Learn-Then-Test and returns a valid set inside $\{\lambda : R(\lambda) \le \alpha\}$ with family-wise error control,
\begin{equation}
\Pr\bigl(\,\exists\, \lambda \in \text{valid set} : R(\lambda) > \alpha\,\bigr) \;\le\; \delta.
\end{equation}
Two properties matter downstream. The guarantee holds for any loss distribution and any selection statistic, so the quality of the grid decides only which configurations pass. The statement is also simultaneous over the valid set, which is what makes the deployment rule of Appendix~\ref{app:deploy-simult} free.

\subsection{The exact binomial \texorpdfstring{$p$}{p}-value}
\label{app:pvalue}

For $H_0 : R(\lambda) > \alpha$ with Bernoulli losses, the violation count $S = \sum_i L_i(\lambda)$ satisfies $S \sim \mathrm{Bin}(n, R)$. The $p$-value
\begin{equation}
p(\lambda) = \Pr_{\mathrm{Bin}(n, \alpha)}\bigl(\,S \le V(\lambda)\,\bigr), \qquad V(\lambda) = \text{observed violations},
\end{equation}
is super-uniform under $H_0$ by stochastic ordering. $\mathrm{Bin}(n, R)$ with $R > \alpha$ stochastically dominates $\mathrm{Bin}(n, \alpha)$, so small counts are rarer under every null than at the boundary $R = \alpha$. Rejecting at level $\delta$ therefore establishes $R \le \alpha$ with error probability at most $\delta$.

\subsection{Holm step-down under arbitrary dependence}
\label{app:holm}

Order $p_{(1)} \le \cdots \le p_{(m)}$ and reject while $p_{(i)} \le \delta/(m - i + 1)$, stopping at the first failure. \citet{holm1979simple} shows that this controls the family-wise error rate at $\delta$ under \textbf{arbitrary} joint dependence of the $p$-values. Our setting needs exactly this property, because every configuration is scored on the \emph{same} calibration prompts, which induces strong positive dependence. Holm rejects a superset of what Bonferroni rejects, since each ordered threshold $\delta/(m - i + 1)$ is at least $\delta/m$, and we keep Bonferroni as the conservative floor. Appendix~\ref{app:grids-frontier} quantifies the gap between the two.

\subsection{Fixed sequence as secondary}
\label{app:fixedseq}

The fixed-sequence procedure walks the configurations in the rank order of the design thresholds and tests each at full $\delta$ until the first failure. Its validity needs no monotonicity of the risk, because an ordering independent of the calibration losses suffices, and monotonicity affects only its power. We report it as secondary and never mix selections across procedures at one $\alpha$. It is not primary because measured risk is \textbf{non-monotone} in the design thresholds. The configurations with accept threshold $0.99$ in Appendix~\ref{app:grids-nonmonotone} are slower \emph{and} riskier than the reference, so any such walk can stop early on an out-of-order configuration, whereas Holm needs no order.

\subsection{Simultaneity of the deployment rule}
\label{app:deploy-simult}

The deployed configuration is $\lambda^\star = \operatorname{arg\,max}_{\lambda\ \text{valid}} \text{measured TPF}$. TPF is a point estimate that never enters any $p$-value, and the family-wise statement of Appendix~\ref{app:setup} covers every valid configuration \textbf{simultaneously}. The post-hoc choice of the fastest configuration therefore costs nothing,
\begin{equation}
\Pr\bigl(R(\lambda^\star) > \alpha\bigr) \;\le\; \Pr\bigl(\exists\, \lambda \in \text{valid set} : R(\lambda) > \alpha\bigr) \;\le\; \delta .
\end{equation}

\subsection{Error levels within and across grids}
\label{app:joint-delta}

The guarantee of each grid holds at confidence $1 - \delta$ with $\delta = 0.10$, and the paper presents its four primary block-diffusion grids individually and never pooled. When the conjunction is wanted, Bonferroni over the four grids governs, each grid running at $\delta/4 = 0.025$. Under that correction the SDAR math configuration deployed at $\alpha = 0.10$ is still valid, with $p = 8.5\times10^{-4}$ under the Holm threshold $0.025/6 \approx 0.0042$ at its rank.

\textbf{The smallest budget with a gain.} Section~\ref{sec:exp-unlock} reads, for each grid, the smallest budget $\hat\alpha$ on the budget grid at which a configuration faster than the reference is valid under Holm. No guarantee that holds jointly along the $\alpha$ grid is claimed, and none is needed for that one statement.

Let $F$ be the number of faster configurations, five in each grid, and $\alpha_0$ the smallest grid budget at or above the smallest true risk among them. If $\hat\alpha < \alpha_0$, then every faster configuration is a true null at $\hat\alpha$. Holm rejects a prefix of the sorted $p$-values, so for any faster configuration to be valid the first faster configuration in that order must clear its threshold, and since at most the reference and the slower configurations precede it, that threshold is at most $\delta/F$. Each $p$-value decreases in $\alpha$, so the event is contained in $\{\min_{\text{faster}} p_\lambda(\alpha_0^-) \le \delta/F\}$ at the single grid budget $\alpha_0^-$ just below $\alpha_0$, where all $F$ of those $p$-values are super-uniform, and the union bound gives probability at most $\delta$. Hence $\Pr(\hat\alpha \ge \alpha_0) \ge 1 - \delta$ for each grid and $1 - 2\delta$ for the two math grids jointly. The statement concerns the budget and not the configuration valid at $\hat\alpha$, so deployments are reported at the three budgets of Table~\ref{tab:unlock}, and deploying the reference is a non-rejection that carries no statement about its task.

\subsection{Empirical validity at three layers}
\label{app:validity}

The guarantee is checked at three layers, each catching what the previous one cannot.
\textbf{Theorem.} Super-uniform $p$-values combined with Holm control the family-wise error under arbitrary dependence, as Appendices~\ref{app:pvalue} and~\ref{app:holm} show.

\textbf{\boldmath Synthetic Monte Carlo with known risk.} Five configurations carry true risks $\{0.01, 0.02, 0.07, 0.07, 0.07\}$ at $\alpha = 0.05$, two below the budget and three above it, with $n = 800$ and $4{,}000$ trials for each procedure. For Holm, the losses of all five configurations are drawn from a \emph{shared} prompt-level uniform, which induces the positive dependence across configurations of the real setting, and the empirical family-wise error rate is $0.0003$. Fixed sequence and Bonferroni, run on independent draws, read $0$.

\textbf{Out-of-sample coverage on real data.} Twelve grids, $K = 1{,}000$ random calibration and test splits each, budgets $\alpha \in \{0.05, 0.10, 0.15, 0.20\}$ and two to five calibration sizes give $220$ combinations, summarized in Figure~\ref{fig:coverage}. The twelve grids are the six block-diffusion grids of Appendix~\ref{app:grids}, a LLaDA2 math grid on $32$ prompts of each subtask, the speculative-decoding grids of both draft pairs on GSM8K and on HumanEval, and the quantization grid. The minimum coverage is $0.904$, the maximum empirical family-wise error rate is $0.096$, and no combination exceeds the nominal level. Measured against each valid configuration's risk pooled over all $n$ prompts rather than its test-half estimate, a proxy that the calibration half also informs, the same $220$ combinations give a maximum family-wise error rate of $0.014$, and no combination exceeds $\delta$ under either measure. This layer resamples a finite benchmark under exchangeability, so it complements the theorem rather than replacing it.

\section{Measurement Protocols}
\label{app:prereg}

Each measurement of Section~\ref{sec:cmech} is read against a control. Live predecessor finalization is compared with the engine's own traces during serving, self-distillation with the base checkpoint, and the static skip rule with a learned participation gate and an unconstrained oracle over the same features. Appendix~\ref{app:mec} proves the commit invariant and checks it on the engine's traces.

\subsection{Live predecessor finalization}
\label{app:prereg-deltar}

\textbf{Protocol.} The measured quantity is the live flip rate. We call a committed position exposed when, at its step, the committing block has at least one older active block with at least one masked position. The flip rate is the probability over exposed positions that the argmax under the predecessor-finalized counterfactual differs from the token actually committed. A replay counts only when its recomputed argmax reproduces the committed token on at least $99\%$ of the exposed positions, and every main replay passes this check at $99.39$ to $99.70\%$. The main replay covers the requests that finish within the generation cap. Every live rate carries a Wilson $95\%$ interval and a request-level cluster bootstrap $95\%$ interval over $2{,}000$ resamples, since positions within a request correlate.

\textbf{Live and offline readings.} On the engine's own traces the flip rate is $2.88\%$ of $117{,}731$ exposed committed positions on SDAR, with cluster interval $[2.65, 3.12]$, and $3.17\%$ of $137{,}540$ on LLaDA2, with cluster interval $[2.93, 3.43]$. Adding the early decode steps of the $50$ SDAR requests that reached the generation cap gives $2.90\%$. An offline synthetic factorial reads $9.63\%$ $[8.24, 11.23]$ under the block-causal mask and $10.91\%$ $[9.4, 12.6]$ under a bidirectional mask at $n = 1{,}500$, so the live rate is about a third of every synthetic reading in both families, and no live interval reaches a synthetic one. The synthetic settings force a joint masking corner that serving rarely visits. Live, $76$ to $77\%$ of exposed commits face a predecessor at most $25\%$ masked, and that corner flips under $4\%$. Over all committed tokens, the identity-flip rate weighted by serving is $0.888\%$ on SDAR and $0.876\%$ on LLaDA2.

\textbf{Entropy of the deeper blocks.} The companion measurement asks what the masked slots of the deeper blocks lack while they wait. On the first $100$ GSM8K requests of the same SDAR reference-configuration trace, $99$ of which pass the reconstruction check, every decode step with at least two active blocks holding masked generation slots, $6{,}593$ steps in all, was forwarded again unchanged and, for each deeper block, with the masked slots of every older active block finalized to their eventual tokens. The predictive entropy of every masked slot was read from the mask-suppressed softmax that the engine commits from, and the argmax of the unchanged forward reproduces the committed token at $99.64\%$ of exposed positions.

The statistic is the share of steps in which the slot-weighted mean entropy of the deeper blocks exceeds that of the first active block. It is $96.7\%$ in the served state, with $95\%$ request-cluster bootstrap interval $[95.8, 97.6]$ and $3.02$ against $1.53$ nats on average over steps, and $67.3\%$ with predecessors finalized, with interval $[65.3, 69.3]$ and $2.10$ nats. Finalizing the predecessors removes about three fifths of the excess entropy of the deeper blocks. By block rank, the share of masked slots at or above the accept threshold of $0.95$ of the reference configuration is $11.3\%$ in the first active block, $2.6\%$ in the second and $3.0\%$ in the third in the served state, and $15.2\%$ and $21.4\%$ in the second and third once their predecessors are finalized.

The committing forward responds to predecessor state at the masked slots, whereas the positions it commits keep their identity under finalization in $97\%$ of exposed cases. About three fifths of the excess uncertainty of a deeper block is predecessor text that the engine supplies by generating it.

\subsection{Self-distillation on engine-decoded targets}
\label{app:prereg-ctrain}

\textbf{Protocol.} The training targets follow the engine's own decoding. They come from $1{,}000$ GSM8K prompts disjoint from the $n = 256$ evaluation prompts, with at most $12$ blocks of $32$ tokens, decoded iteratively within each block with the engine's confidence rule and with blocks strictly sequential. The rule decodes greedily at threshold $0.95$ and falls back to the top-1 token when no token clears it. These completions reach $0.88$ to $0.89$ accuracy. Training runs for $400$ single-block steps drawn from all $7{,}130$ blocks of these completions at learning rate $2\times10^{-5}$, and engine accuracy and TPF are read against the base checkpoint.

\textbf{Result.} Table~\ref{tab:ctrain} in Section~\ref{sec:cmech-train} lists both checkpoints at the three admission thresholds. The distilled checkpoint gains $+5.2$, $+5.8$ and $+5.8\%$ TPF with accuracy changes of $+0.4$, $+0.8$ and $+0.0$ points, none of them significant, with $p \ge 0.84$.

\textbf{Learn, then test.} The distilled checkpoint was served at the eight threshold settings of the SDAR math grid on the same engine build and tested beside the eight hand-picked configurations as one grid of $m = 16$ hypotheses. Its training pool, GSM8K rows $256$ to $1{,}255$, overlaps the GSM8K prompts of the math set, so the test uses the MATH-500 half ($n = 500$), which neither training nor selection touched, with speed and risk measured against the reference of the grid on the same prompts. Table~\ref{tab:learncert} lists every configuration. The loosest threshold setting of the checkpoint, acc85/semi70, runs faster than every hand-picked configuration, and at every budget $\alpha \ge 0.18$, Redline, run on all sixteen configurations, deploys it at $+46.7\%$ TPF, with $\hat R = 0.146$ ($73$ of $500$), against $+39.5\%$ for the best hand-picked configuration.

\begin{table}[t]
\caption{Redline on MATH-500 ($n = 500$) over the eight hand-picked SDAR math configurations and the distilled checkpoint at the same threshold settings ($m = 16$). TPF, forwards and tokens of each request, joint risk $\hat R$ and $p$-values. Bold marks configurations valid under Holm, and underlining the deployed one.}
\label{tab:learncert}
\centering\footnotesize
\setlength{\tabcolsep}{2pt}
\begin{tabular*}{\textwidth}{@{\extracolsep{\fill}}lrrrrcccc@{}}
\toprule
 &  &  &  &  & \multicolumn{4}{c}{\bfseries\boldmath $p$-value at budget $\alpha$} \\
\cmidrule(l){6-9}
\bfseries\boldmath configuration & \bfseries\boldmath TPF & \bfseries\boldmath forwards & \bfseries\boldmath tokens & \bfseries\boldmath $\hat R$ ($k/n$) & \bfseries\boldmath $0.05$ & \bfseries\boldmath $0.10$ & \bfseries\boldmath $0.15$ & \bfseries\boldmath $0.20$ \\
\midrule
\grouprow{9}{Hand-picked configurations of the SDAR math grid} \\
acc85/semi70 & 5.829 & 147.2 & 858 & 0.106 (53/500) & 1.000 & 0.704 & \underline{\textbf{0.003}} & \textbf{1e-8} \\
acc90/semi70 & 5.490 & 164.5 & 903 & 0.118 (59/500) & 1.000 & 0.919 & \textbf{0.023} & \textbf{8e-7} \\
acc85/semi90 & 5.390 & 161.2 & 869 & 0.102 (51/500) & 1.000 & 0.596 & \textbf{0.001} & \textbf{3e-9} \\
acc90/semi90 & 4.752 & 177.9 & 845 & 0.094 (47/500) & 1.000 & 0.361 & \textbf{1e-4} & \textbf{9e-11} \\
acc95/semi70 & 4.724 & 195.7 & 924 & 0.114 (57/500) & 1.000 & 0.867 & \textbf{0.012} & \textbf{2e-7} \\
acc95/semi90 (reference) & 4.180 & 210.5 & 880 & 0.000 (0/500) & \underline{\textbf{7e-12}} & \underline{\textbf{1e-23}} & \textbf{5e-36} & \textbf{4e-49} \\
acc99/semi70 & 3.273 & 256.0 & 838 & 0.100 (50/500) & 1.000 & 0.538 & \textbf{7e-4} & \textbf{1e-9} \\
acc99/semi90 & 3.189 & 259.6 & 828 & 0.060 (30/500) & 0.869 & \textbf{0.001} & \textbf{3e-10} & \textbf{6e-19} \\
\grouprow{9}{Distilled checkpoint at the same eight threshold settings} \\
acc85/semi70 & 6.133 & 153.9 & 944 & 0.146 (73/500) & 1.000 & 1.000 & 0.431 & \underline{\textbf{0.001}} \\
acc85/semi90 & 5.463 & 160.2 & 875 & 0.148 (74/500) & 1.000 & 1.000 & 0.481 & \textbf{0.002} \\
acc90/semi70 & 5.381 & 162.0 & 871 & 0.122 (61/500) & 1.000 & 0.954 & 0.043 & \textbf{3e-6} \\
acc90/semi90 & 4.945 & 173.8 & 860 & 0.092 (46/500) & 1.000 & 0.306 & \textbf{8e-5} & \textbf{4e-11} \\
acc95/semi70 & 4.826 & 181.6 & 876 & 0.110 (55/500) & 1.000 & 0.796 & \textbf{0.006} & \textbf{5e-8} \\
acc95/semi90 & 4.153 & 202.0 & 839 & 0.070 (35/500) & 0.980 & 0.012 & \textbf{3e-8} & \textbf{4e-16} \\
acc99/semi70 & 3.649 & 246.6 & 900 & 0.090 (45/500) & 1.000 & 0.255 & \textbf{4e-5} & \textbf{1e-11} \\
acc99/semi90 & 3.250 & 256.9 & 835 & 0.056 (28/500) & 0.768 & \textbf{3e-4} & \textbf{3e-11} & \textbf{4e-20} \\
\bottomrule
\end{tabular*}
\end{table}

\subsection{Participation schedules}
\label{app:prereg-calloc}

\textbf{Cost model.} In both families, a participation schedule pays off when it cuts the FLOPs of each token without paying the saving back in extra steps. Wall-clock cost is modeled in token-equivalent units as $\text{steps} \cdot \rho + \text{FLOP}$, where $\rho$ is the fixed overhead of each forward divided by the compute of each token. At $\rho \to 0$, the compute-bound regime that a throughput claim targets, the speedup reduces to the FLOP saving, and at $\rho \to \infty$ to the ratio of step counts. The analysis sweeps the whole range without a GPU.

\textbf{Replay.} The schedules are replayed on the engine's decode traces with the committed tokens of every block held fixed, so that deferring the forward of a block changes only the step count. The static skip rule skips a fully masked block while an older block is still active. The learned participation gate scores the same causal pre-forward features, and the unconstrained oracle removes every zero-commit forward. At $\rho \to 0$ the static rule saves $36$ to $37\%$ of FLOPs in both families, the oracle lies at most $1.3$ points above it, and the learned gate matches the static rule to within $0.02$ points, as Figure~\ref{fig:probes}c shows.

\section{The Maximal-Exact-Commit Invariant}
\label{app:mec}

\subsection{Statement}
\label{app:mec-statement}

\textbf{Proposition (maximal exact commit under default commit semantics).} \emph{Under the engine's default commit rule, the commit sweep of every decode step transfers to the cache, within that step, every active block that is fully resolved, commit-ready and preceded only by committed blocks. In the serving configurations deployed for both model families, a fully resolved block is always commit-ready when the sweep reaches it, so the exact-committable but uncommitted block mass is zero by construction.}

A block-step is \emph{exact-committable but uncommitted} when a block is active, has no masked position and has no uncommitted predecessor, that is, a block the engine could commit at no risk yet does not. Under in-order commit semantics this is the right notion of committability. A complete block behind an uncommitted predecessor cannot be committed, because the cache must stay contiguous, and such steps are counted separately as legitimate in-order waits.

\subsection{Proof}
\label{app:mec-proof}

The proof has three parts, each a property of the serving engine.

\textbf{(1) The commit sweep is exhaustive within a step.} The sweep visits the block buffer from left to right and commits a block exactly when it is active, complete, commit-ready and its predecessor is committed. Blocks are examined in strictly increasing order, and removing the first block shifts the indices so that no block is skipped or examined twice. The predecessor condition is monotone within a pass, because the transitions into and within the cache are one-way. Completeness and commit readiness cannot change during the sweep, since all sampler writes and state updates precede it. A freshly committed block satisfies its successor's predecessor condition in the same pass, and because each predecessor is examined first and its transitions are one-way, its status is final when its successor is examined.

Chains of ready blocks therefore commit in a single step, which matches the measured distribution of multi-block commits ($295$, $32$ and $10$ steps committing two, three and four blocks at once on an SDAR trace with a four-block buffer). Committed blocks form a prefix of the buffer under in-order commit, so removing the first block is benign. After every sweep, no block remains that is active, complete, commit-ready and preceded only by committed blocks.

\textbf{\boldmath (2) Complete implies commit-ready for both deployed samplers.} The scheduler sets a block's commit readiness from a state map when the sampler provides one, and otherwise marks every complete active block commit-ready. The SDAR sampler provides no state map. The LLaDA2 samplers that do provide one are selected only by the in-place revision mode of LLaDA2.1 and by the token-merge decoding of DMax, and no deployed or measured configuration selects either. Every measured and tested configuration runs the plain LLaDA2 sampler, which provides no state map, so the executed path contains no deferral clause. The measured zero of Appendix~\ref{app:mec-validation} also confirms the plain path, since the quiescence check of the revision sampler defers essentially every block completion by one step.

\textbf{(3) Snapshots observe the state after the sweep.} Each engine step runs the sweep before it returns, and the generation loop records the trajectory snapshot afterwards, with nothing changing the request state in between. A preempted request re-enters through steps that re-encode its prompt, which the check excludes, and the sweep runs in every step, so the invariant holds at every recorded decode snapshot.

Parts (1) to (3) together give the proposition. \hfill$\blacksquare$

\subsection{Validation on decode traces}
\label{app:mec-validation}

A check counts exact-committable but uncommitted block-steps over four decode traces, two families at buffer depths $2$ and $4$ on math with $128$ prompts in each benchmark, and finds \textbf{none among $281{,}005$ decode block-steps} ($61{,}587$, $66{,}811$, $74{,}550$ and $78{,}057$). Its predicate ignores commit readiness altogether and counts blocks that are active, complete and preceded only by committed blocks, so the measured zero is stronger than the sweep invariant. It also rules out any commit-ready deferral and confirms the snapshot ordering of part (3).

\subsection{Scope}
\label{app:mec-scope}

The proposition covers default commit semantics. The revision-capable sampler modes, in-place revision in LLaDA2.1 and token merging in DMax, admit a one-step quiescence deferral and lie outside it, since in those modes the content of a block without masks can still change.

\section{Full Grids and Selector Comparisons}
\label{app:grids}

\subsection{Grid tables}
\label{app:grids-tables}

Tables~\ref{tab:grid_llada2_math} to~\ref{tab:grid_llada2_code_L256} report, for every configuration of every grid, the measured TPF, the forwards and tokens of each request, the empirical joint risk $\hat R = k/n$ (reference correct and configuration wrong), the exact binomial $p$-value at each printed budget, whether the configuration is valid under Holm, and the deployed configuration.

\textbf{Intervals for every deployed configuration.} Table~\ref{tab:intervals} attaches intervals to every deployed configuration other than the reference. The speed gain and the net accuracy change carry $95\%$ paired-bootstrap percentile intervals over prompts with $B = 10{,}000$, the same resample weighting the deployed configuration and the reference, and the empirical joint risk carries its two-sided $95\%$ Clopper--Pearson interval and the one-sided $90\%$ upper bound that a single test at $1 - \delta$ would report. Across the $23$ such entries at the four printed budgets, all $20$ speed intervals lie above zero and every upper bound lies below its budget. The net intervals of the Qwen speculative-decoding deployments and of the quantized Llama include zero, so those deployments are not shown to cost accuracy at $n = 512$. The quantization gain is a measured property of the precision and carries no sampling error. The intervals are conditional on the configuration that the full-sample procedure selected, and Appendix~\ref{app:heldout-deploy} re-measures the deployed configurations on held-out prompts.

\begin{table}[t]
\caption{Intervals for every configuration that Redline deploys, other than the reference. Gain and net accuracy change with $95\%$ paired-bootstrap intervals over prompts, and joint risk $\hat R$ with its $95\%$ Clopper--Pearson interval and one-sided $90\%$ upper bound. Budgets that deploy the same configuration share a row.}
\label{tab:intervals}
\centering\footnotesize
\setlength{\tabcolsep}{1.6pt}
\begin{tabular*}{\textwidth}{@{\extracolsep{\fill}}l c rc rc r c c@{}}
\toprule
 &  & \multicolumn{2}{c}{\bfseries\boldmath gain} & \multicolumn{2}{c}{\bfseries\boldmath net accuracy} & \multicolumn{3}{c}{\bfseries\boldmath joint risk} \\
\cmidrule(lr){3-4}\cmidrule(lr){5-6}\cmidrule(l){7-9}
\bfseries\boldmath grid & \bfseries\boldmath $\alpha$ & \bfseries\boldmath value & \bfseries\boldmath $95\%$ interval & \bfseries\boldmath value & \bfseries\boldmath $95\%$ interval & \bfseries\boldmath $\hat R$ & \bfseries\boldmath $95\%$ interval & \bfseries\boldmath \begin{tabular}[b]{@{}c@{}}$90\%$\\bound\end{tabular} \\
\midrule
\grouprow{9}{Block-diffusion serving, TPF gain (\%)} \\
LLaDA2 math & $\ge 0.10$ & $+37.5$ & $[34.0, 41.1]$ & $-4.0$ & $[-5.9, -2.0]$ & $0.072$ & $[0.057, 0.090]$ & $0.084$ \\
LLaDA2 math, large grid & $0.10$ & $+65.4$ & $[60.8, 70.2]$ & $-3.4$ & $[-5.4, -1.4]$ & $0.073$ & $[0.058, 0.091]$ & $0.085$ \\
LLaDA2 math, large grid & $\ge 0.15$ & $+72.0$ & $[67.7, 76.3]$ & $-5.9$ & $[-8.2, -3.8]$ & $0.098$ & $[0.080, 0.118]$ & $0.111$ \\
LLaDA2 code & $0.15$ & $+33.7$ & $[26.8, 41.1]$ & $-7.0$ & $[-10.1, -3.9]$ & $0.109$ & $[0.084, 0.138]$ & $0.128$ \\
LLaDA2 code & $0.20$ & $+42.4$ & $[33.1, 52.2]$ & $-13.1$ & $[-16.6, -9.8]$ & $0.157$ & $[0.127, 0.190]$ & $0.179$ \\
SDAR math & $0.10$ & $+13.1$ & $[7.1, 19.6]$ & $-2.5$ & $[-4.6, -0.4]$ & $0.071$ & $[0.056, 0.089]$ & $0.083$ \\
SDAR math & $\ge 0.15$ & $+36.6$ & $[29.2, 44.4]$ & $-4.6$ & $[-7.1, -2.3]$ & $0.105$ & $[0.087, 0.125]$ & $0.118$ \\
SDAR code & $0.15$ & $+25.7$ & $[17.6, 38.9]$ & $-5.0$ & $[-7.5, -2.4]$ & $0.087$ & $[0.067, 0.111]$ & $0.103$ \\
SDAR code & $0.20$ & $+59.7$ & $[38.5, 88.2]$ & $-14.0$ & $[-17.5, -10.7]$ & $0.178$ & $[0.149, 0.209]$ & $0.198$ \\
\grouprow{9}{Speculative decoding, gain in accepted tokens for each target forward (\%)} \\
Llama GSM8K & $0.10$ & $+13.8$ & $[13.2, 14.5]$ & $-3.3$ & $[-6.2, -0.6]$ & $0.072$ & $[0.051, 0.098]$ & $0.089$ \\
Llama GSM8K & $\ge 0.15$ & $+16.8$ & $[16.1, 17.5]$ & $-9.0$ & $[-12.3, -5.7]$ & $0.123$ & $[0.096, 0.155]$ & $0.144$ \\
Qwen GSM8K & $0.05$ & $+15.0$ & $[14.3, 15.7]$ & $+0.4$ & $[-2.0, +2.7]$ & $0.033$ & $[0.019, 0.053]$ & $0.046$ \\
Qwen GSM8K & $\ge 0.10$ & $+18.1$ & $[17.4, 18.9]$ & $-1.6$ & $[-4.3, +1.2]$ & $0.057$ & $[0.038, 0.080]$ & $0.072$ \\
\grouprow{9}{Weight quantization, reduction in weight memory} \\
Llama GSM8K & $\ge 0.10$ & $2.81\times$ & measured & $-1.4$ & $[-4.3, +1.6]$ & $0.064$ & $[0.045, 0.089]$ & $0.081$ \\
\bottomrule
\end{tabular*}
\end{table}

\subsection{Non-monotone risk}
\label{app:grids-nonmonotone}

In all four main grids, risk is \textbf{non-monotone in the accept threshold}. The configurations with accept threshold $0.99$ are \emph{slower} than the reference yet carry nonzero reference-relative risk. On SDAR math, acc99/semi70 runs at TPF $3.09$ against $3.85$ for the reference with $\hat R = 0.084$ ($85$ of $1{,}012$), and on LLaDA2 code, acc99/semi90 runs at $3.69$ against $5.32$ with $\hat R = 0.035$. A practitioner hand-tuning conservative thresholds therefore gets a configuration both slower and riskier than the default, with no signal saying so. Holm needs no order across the grid, which is why it is primary and why the fixed-sequence walk, which needs one, is secondary.

\subsection{Deployed gain across budgets, and Holm against Bonferroni}
\label{app:grids-frontier}

\begin{figure}[t]
\centering
\includegraphics[width=\linewidth]{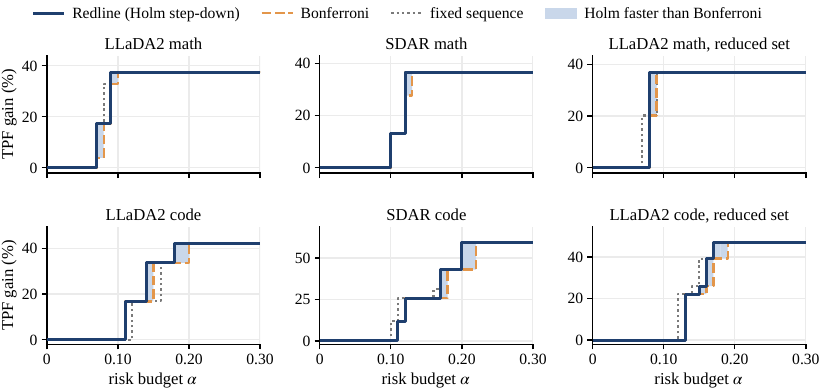}
\caption{Deployed TPF gain across risk budgets for the six block-diffusion grids under Holm, Bonferroni and fixed-sequence testing. Rows are tasks, and the right column holds the two grids on a reduced prompt set. Shading marks budgets where Holm deploys a faster configuration than Bonferroni.}
\label{fig:frontier}
\end{figure}

\textbf{Holm against Bonferroni.} At identical $(\alpha, \delta)$, Holm's valid set contains Bonferroni's by construction. Holm raises the deployed gain over Bonferroni's at grid-specific budgets between $0.07$ and $0.19$, by up to $13$ to $17$ points of TPF gain across the four LLaDA2 grids, with exact ties elsewhere, as Figure~\ref{fig:frontier} shows. On SDAR the gains are $+8.9$ points at $\alpha = 0.12$ on math, and $+17.3$ points at $\alpha = 0.17$ and $+16.6$ points at $\alpha = 0.20$ to $0.21$ on code. On the Fast-dLLM-v2 grid of Appendix~\ref{app:grids-absorption} the same gap reappears at $\alpha = 0.14$ ($+9.7$ points) and $\alpha = 0.19$ ($+22.4$ points), and on the grid of $50$ configurations of Appendix~\ref{app:grids-biggrid} at three budgets in each of its two larger nested grids.

Each gain arises when a configuration's $p$-value lands between Bonferroni's $\delta/m$ and Holm's threshold $\delta/(m - i + 1)$ at its rank $i$, most visibly at Holm's last step at full $\delta$. On LLaDA2 math at $\alpha = 0.07$, its smallest budget with a gain, the acc90/semi90 configuration with $\hat R = 0.053$ has $p = 0.0190$, which fails Bonferroni's $0.0143$ and passes Holm's $0.025$. Holm therefore deploys this configuration at TPF $5.17$ ($+17.4\%$), where Bonferroni deploys acc95/semi70 at $4.57$ ($+3.9\%$).

\textbf{\boldmath Sensitivity to $\delta$.} Repeating the test on every grid at $\delta \in \{0.10, 0.05, 0.01\}$ leaves the math configurations deployed at $\alpha = 0.10$ unchanged in both families, LLaDA2 acc85/semi70 at $+37.5\%$ and SDAR acc90/semi90 at $+13.1\%$, down to $\delta = 0.01$. Across the full $\alpha$ grid only $20$ of the $360$ evaluations over grids, budgets and the three values of $\delta$ change the deployed configuration, all at budgets where a configuration sits at the rejection boundary, as expected of an exact test.

\textbf{Calibration size and the task ordering.} At $\alpha = 0.10$ Holm's step passes a configuration only when its empirical risk $\hat R$ clears the budget by a finite-sample margin, about $2.1$pp at $n = 1{,}012$ and $2.8$ to $3.0$pp at $n = 542$. At its observed rate but with $n = 1{,}012$ prompts, the LLaDA2 code configuration would be valid at $\alpha = 0.10$ ($p = 0.005$, deployed gain $+16.8\%$), and the smallest budgets with a gain would keep their order, LLaDA2 math at $0.07$ against code at $0.10$.

The ordering also holds at equal prompt counts. Table~\ref{tab:subsample} draws $1{,}000$ subsets of the $1{,}012$ math prompts at the size of the code set without replacement and re-runs Redline on each over the fine budget grid. The smallest LLaDA2 math budget with a gain has median $0.08$ and lies below the code value of $0.11$ in every draw, and the SDAR one has median $0.10$ and lies at or below $0.11$ in $99.1\%$ of draws.

In both families, each of the five configurations faster than the reference fails on a larger share of the reference's correct answers on code than on math, $10.5$ to $21.9\%$ against $5.9$ to $8.4\%$ for LLaDA2 and $17.8$ to $38.9\%$ against $9.2$ to $13.5\%$ for SDAR, a ratio between $1.7$ and $2.9$ at every configuration. Table~\ref{tab:samplesize} gives the calibration size from which a configuration one or two points under the budget passes with power at least $0.8$ at every larger $n$, at Holm's first step, the most demanding one.

\begin{table}[t]
\caption{Smallest math budget with a gain, with the math set subsampled to the code $n$, over $1{,}000$ uniform or benchmark-stratified draws for each family (Redline, $\delta = 0.10$, fine budget grid). Median, interquartile range, shares of draws under the listed budgets, and the draws at each budget.}
\label{tab:subsample}
\centering\footnotesize\setlength{\tabcolsep}{1.5pt}
\begin{tabular*}{\textwidth}{@{\extracolsep{\fill}}llc cc ccc l@{}}
\toprule
\bfseries\boldmath family & \bfseries\boldmath draw & \bfseries\boldmath $n$ & \bfseries\boldmath median & \bfseries\boldmath IQR & \bfseries\boldmath $\le 0.10$ & \bfseries\boldmath $< 0.11$ & \bfseries\boldmath $\le 0.11$ & \bfseries\boldmath \begin{tabular}[b]{@{}l@{}}draws by smallest budget\\with a gain, $0.05$ to $0.12$\end{tabular} \\
\midrule
LLaDA2 math & uniform & 542 & 0.08 & [0.07, 0.08] & 1.000 & 1.000 & 1.000 & 0, 52, 334, 543, 70, 1, 0, 0 \\
LLaDA2 math & stratified & 542 & 0.08 & [0.07, 0.08] & 1.000 & 1.000 & 1.000 & 1, 36, 324, 551, 87, 1, 0, 0 \\
SDAR math & uniform & 664 & 0.10 & [0.10, 0.11] & 0.739 & 0.739 & 0.991 & 0, 0, 0, 14, 224, 501, 252, 9 \\
SDAR math & stratified & 664 & 0.10 & [0.09, 0.11] & 0.741 & 0.741 & 0.994 & 0, 0, 0, 13, 252, 476, 253, 6 \\
\bottomrule
\end{tabular*}
\end{table}

\begin{table}[t]
\caption{Calibration size for the exact binomial test ($\delta = 0.10$). (a) Prompts from which a configuration of true risk $R$ passes with probability at least $0.8$ at Holm's first step $\delta/m$. (b) The largest empirical risk that passes at $\alpha = 0.10$ with $n$ prompts.}
\label{tab:samplesize}
\centering\footnotesize
\setlength{\tabcolsep}{2pt}
\begin{minipage}[t]{0.54\textwidth}\centering
\textit{(a) Prompts for power $0.8$}\\[2pt]
\begin{tabular*}{\linewidth}{@{\extracolsep{\fill}}cc rrr@{}}
\toprule
 &  & \multicolumn{3}{c}{\bfseries\boldmath grids of $m$ configurations} \\
\cmidrule(l){3-5}
\bfseries\boldmath $\alpha$ & \bfseries\boldmath true $R$ & \bfseries\boldmath $m = 8$ & \bfseries\boldmath $m = 16$ & \bfseries\boldmath $m = 50$ \\
\midrule
$0.02$ & $0.01$ & $1{,}623$ & $1{,}828$ & $2{,}274$ \\
$0.05$ & $0.04$ & $4{,}320$ & $5{,}048$ & $6{,}217$ \\
$0.05$ & $0.03$ & $1{,}008$ & $1{,}189$ & $1{,}444$ \\
$0.10$ & $0.09$ & $8{,}407$ & $9{,}842$ & $12{,}179$ \\
$0.10$ & $0.08$ & $2{,}066$ & $2{,}406$ & $2{,}966$ \\
\bottomrule
\end{tabular*}
\end{minipage}\hfill
\begin{minipage}[t]{0.42\textwidth}\centering
\textit{(b) Margin at $\alpha = 0.10$}\\[2pt]
\begin{tabular*}{\linewidth}{@{\extracolsep{\fill}}r rrr@{}}
\toprule
 & \multicolumn{3}{c}{\bfseries\boldmath largest $\hat R$ (\%)} \\
\cmidrule(l){2-4}
\bfseries\boldmath $n$ & \bfseries\boldmath $\delta/8$ & \bfseries\boldmath $\delta/50$ & \bfseries\boldmath $\delta$ \\
\midrule
$542$ & $7.0$ & $6.3$ & $8.1$ \\
$664$ & $7.4$ & $6.6$ & $8.4$ \\
$1{,}012$ & $7.8$ & $7.3$ & $8.7$ \\
$2{,}000$ & $8.5$ & $8.1$ & $9.1$ \\
$5{,}000$ & $9.0$ & $8.8$ & $9.4$ \\
\bottomrule
\end{tabular*}
\end{minipage}
\end{table}

\subsection{Grids on a reduced prompt set}
\label{app:grids-l256}

Tables~\ref{tab:grid_llada2_math_L256} and~\ref{tab:grid_llada2_code_L256} report two further LLaDA2 grids on a reduced prompt set of the same structure, at most $256$ prompts in each subtask, so $256$ GSM8K and $256$ MATH-500 prompts for math ($n = 512$) and $n = 420$ for code. Their frontiers form the right column of Figure~\ref{fig:frontier}, where Holm's gain over Bonferroni peaks at $+16.7$ points at $\alpha = 0.08$ on math and $+13.2$ points at $\alpha = 0.16$ on code. The main text uses the full grids throughout.

\subsection{Forwards and verbosity}
\label{app:grids-nfe}

TPF conflates fewer forwards with longer outputs, so the grid tables co-report the forwards and tokens of each request for every configuration. The SDAR math configuration deployed at $\alpha = 0.10$ (acc90/semi90) cuts compute cleanly, with the forwards of each request falling from $147.4$ to $123.9$ ($-15.9\%$) and tokens by $4.9\%$, so its $+13.1\%$ TPF understates the saving.

\subsection{Oracle gates under matched accounting}
\label{app:grids-oracle}

The oracle of Figure~\ref{fig:probes}c uses the same FLOP-weighted accounting and the same replay as the static rule and the learned gate. We report two oracle notions.
\textbf{Unconstrained ceiling.} Every zero-commit forward is removed at no scheduling cost, the upper bound for any gate that leaves commits unchanged. It saves $37.31\%$ FLOP on LLaDA2 and $38.31\%$ on SDAR, \textbf{\boldmath $+0.98$pp and $+1.30$pp} over the static rule.

\textbf{Schedulable oracle.} It skips exactly the zero-commit forwards, replayed closed loop, and pays for a forced forward whenever every active block would skip. It saves $36.26\%$ and $37.36\%$, \textbf{\boldmath $-0.07$pp and $+0.35$pp} against the static rule, which numerically exceeds it on LLaDA2. Both schedules drop a block's leading zero-commit forwards for free. The static rule still pays for its unskipped zeros, those of the oldest block and those inside a block after its first commit, and the oracle pays for its forced steps, which form the larger total on LLaDA2.

Even with perfect knowledge of every forward, then, no realizable gate clears the static rule by more than $0.35$pp in either family under the accounting that the throughput regime uses.

\subsection{A three-dimensional grid}
\label{app:grids-3d}

The block-add schedule enters as a third tested dimension (accept $\times$ semi $\times$ add, $9$ configurations, $n = 1{,}012$). Redline deploys the same acc85/semi70 configuration at $+37.5\%$ TPF at $\alpha = 0.10$, which reproduces the two-dimensional result exactly, and the add axis is never selected, since both of its directions buy about $4.5$ to $4.7$pp of risk for at most $+0.7\%$ TPF. Redline needs no ordering of the thresholds in three dimensions either.

\subsection{An external engine}
\label{app:grids-absorption}

Every speedup below is reported with its $\alpha$ at $\delta = 0.10$. \textbf{Fast-dLLM-v2} (Qwen, GSM8K, $n = 512$) yields a deployed frontier over a five-configuration grid of accept thresholds, with deployed gains of $+15.3\%$ TPF at $\alpha = 0.10$ ($\hat R = 0.070$), $+25.0\%$ at $\alpha = 0.15$ and $+47.4\%$ at $\alpha = 0.20$.

\begin{table}[t]
\caption{Full grid for LLaDA2 math. TPF, forwards and tokens of each request, joint risk $\hat R$ ($k/n$) and exact binomial $p$-values at four budgets. Bold marks the configurations valid under Holm, and underlining the one Redline deploys.}
\label{tab:grid_llada2_math}
\centering\footnotesize
\setlength{\tabcolsep}{2pt}
\begin{tabular*}{\textwidth}{@{\extracolsep{\fill}}lrrrrcccc@{}}
\toprule
 &  &  &  &  & \multicolumn{4}{c}{\bfseries\boldmath $p$-value at budget $\alpha$} \\
\cmidrule(l){6-9}
\bfseries\boldmath configuration & \bfseries\boldmath TPF & \bfseries\boldmath forwards & \bfseries\boldmath tokens & \bfseries\boldmath $\hat R$ ($k/n$) & \bfseries\boldmath $0.05$ & \bfseries\boldmath $0.10$ & \bfseries\boldmath $0.15$ & \bfseries\boldmath $0.20$ \\
\midrule
acc85/semi70 & 6.050 & 128.8 & 779 & 0.072 (73/1012) & 0.999 & \underline{\textbf{0.001}} & \underline{\textbf{3e-14}} & \underline{\textbf{5e-30}} \\
acc85/semi90 & 5.850 & 134.3 & 786 & 0.066 (67/1012) & 0.990 & \textbf{1e-4} & \textbf{1e-16} & \textbf{4e-33} \\
acc90/semi70 & 5.456 & 144.4 & 788 & 0.064 (65/1012) & 0.981 & \textbf{4e-5} & \textbf{2e-17} & \textbf{3e-34} \\
acc90/semi90 & 5.165 & 149.5 & 772 & 0.053 (54/1012) & 0.718 & \textbf{6e-8} & \textbf{2e-22} & \textbf{7e-41} \\
acc95/semi70 & 4.571 & 171.4 & 784 & 0.051 (52/1012) & 0.616 & \textbf{1e-8} & \textbf{2e-23} & \textbf{3e-42} \\
acc95/semi90 (reference) & 4.401 & 166.3 & 732 & 0.000 (0/1012) & \underline{\textbf{3e-23}} & \textbf{5e-47} & \textbf{4e-72} & \textbf{8e-99} \\
acc99/semi90 & 3.287 & 203.4 & 668 & 0.040 (40/1012) & 0.069 & \textbf{5e-13} & \textbf{3e-30} & \textbf{8e-51} \\
\bottomrule
\end{tabular*}
\end{table}

\begin{table}[t]
\caption{Full grid for LLaDA2 code. TPF, forwards and tokens of each request, joint risk $\hat R$ ($k/n$) and exact binomial $p$-values at four budgets. Bold marks the configurations valid under Holm, and underlining the one Redline deploys.}
\label{tab:grid_llada2_code}
\centering\footnotesize
\setlength{\tabcolsep}{2pt}
\begin{tabular*}{\textwidth}{@{\extracolsep{\fill}}lrrrrcccc@{}}
\toprule
 &  &  &  &  & \multicolumn{4}{c}{\bfseries\boldmath $p$-value at budget $\alpha$} \\
\cmidrule(l){6-9}
\bfseries\boldmath configuration & \bfseries\boldmath TPF & \bfseries\boldmath forwards & \bfseries\boldmath tokens & \bfseries\boldmath $\hat R$ ($k/n$) & \bfseries\boldmath $0.05$ & \bfseries\boldmath $0.10$ & \bfseries\boldmath $0.15$ & \bfseries\boldmath $0.20$ \\
\midrule
acc85/semi70 & 7.566 & 28.9 & 219 & 0.157 (85/542) & 1.000 & 1.000 & 0.697 & \underline{\textbf{0.006}} \\
acc85/semi90 & 7.107 & 28.6 & 203 & 0.109 (59/542) & 1.000 & 0.778 & \underline{\textbf{0.003}} & \textbf{9e-9} \\
acc90/semi70 & 6.793 & 30.4 & 207 & 0.138 (75/542) & 1.000 & 0.998 & 0.245 & \textbf{1e-4} \\
acc90/semi90 & 6.206 & 34.0 & 211 & 0.076 (41/542) & 0.996 & 0.031 & \textbf{1e-7} & \textbf{7e-16} \\
acc95/semi70 & 5.495 & 38.3 & 210 & 0.100 (54/542) & 1.000 & 0.525 & \textbf{4e-4} & \textbf{2e-10} \\
acc95/semi90 (reference) & 5.315 & 38.9 & 207 & 0.000 (0/542) & \underline{\textbf{8e-13}} & \underline{\textbf{2e-25}} & \textbf{6e-39} & \textbf{3e-53} \\
acc99/semi70 & 3.977 & 53.7 & 213 & 0.089 (48/542) & 1.000 & 0.209 & \textbf{1e-5} & \textbf{1e-12} \\
acc99/semi90 & 3.691 & 53.3 & 197 & 0.035 (19/542) & 0.062 & \textbf{9e-9} & \textbf{2e-18} & \textbf{7e-30} \\
\bottomrule
\end{tabular*}
\end{table}

\begin{table}[t]
\caption{Full grid for SDAR math. TPF, forwards and tokens of each request, joint risk $\hat R$ ($k/n$) and exact binomial $p$-values at four budgets. Bold marks the configurations valid under Holm, and underlining the one Redline deploys.}
\label{tab:grid_sdar_math}
\centering\footnotesize
\setlength{\tabcolsep}{2pt}
\begin{tabular*}{\textwidth}{@{\extracolsep{\fill}}lrrrrcccc@{}}
\toprule
 &  &  &  &  & \multicolumn{4}{c}{\bfseries\boldmath $p$-value at budget $\alpha$} \\
\cmidrule(l){6-9}
\bfseries\boldmath configuration & \bfseries\boldmath TPF & \bfseries\boldmath forwards & \bfseries\boldmath tokens & \bfseries\boldmath $\hat R$ ($k/n$) & \bfseries\boldmath $0.05$ & \bfseries\boldmath $0.10$ & \bfseries\boldmath $0.15$ & \bfseries\boldmath $0.20$ \\
\midrule
acc85/semi70 & 5.263 & 102.9 & 542 & 0.105 (106/1012) & 1.000 & 0.714 & \underline{\textbf{2e-5}} & \underline{\textbf{2e-16}} \\
acc90/semi70 & 4.956 & 115.0 & 570 & 0.099 (100/1012) & 1.000 & 0.476 & \textbf{1e-6} & \textbf{2e-18} \\
acc85/semi90 & 4.920 & 112.0 & 551 & 0.093 (94/1012) & 1.000 & 0.244 & \textbf{4e-8} & \textbf{1e-20} \\
acc90/semi90 & 4.361 & 123.9 & 540 & 0.071 (72/1012) & 0.999 & \underline{\textbf{9e-4}} & \textbf{1e-14} & \textbf{2e-30} \\
acc95/semi70 & 4.296 & 136.0 & 584 & 0.083 (84/1012) & 1.000 & 0.037 & \textbf{9e-11} & \textbf{8e-25} \\
acc95/semi90 (reference) & 3.854 & 147.4 & 568 & 0.000 (0/1012) & \underline{\textbf{3e-23}} & \textbf{5e-47} & \textbf{4e-72} & \textbf{8e-99} \\
acc99/semi70 & 3.092 & 178.8 & 553 & 0.084 (85/1012) & 1.000 & 0.047 & \textbf{2e-10} & \textbf{2e-24} \\
acc99/semi90 & 2.891 & 183.8 & 531 & 0.038 (38/1012) & 0.036 & \textbf{6e-14} & \textbf{2e-31} & \textbf{2e-52} \\
\bottomrule
\end{tabular*}
\end{table}

\begin{table}[t]
\caption{Full grid for SDAR code. TPF, forwards and tokens of each request, joint risk $\hat R$ ($k/n$) and exact binomial $p$-values at four budgets. Bold marks the configurations valid under Holm, and underlining the one Redline deploys.}
\label{tab:grid_sdar_code}
\centering\footnotesize
\setlength{\tabcolsep}{2pt}
\begin{tabular*}{\textwidth}{@{\extracolsep{\fill}}lrrrrcccc@{}}
\toprule
 &  &  &  &  & \multicolumn{4}{c}{\bfseries\boldmath $p$-value at budget $\alpha$} \\
\cmidrule(l){6-9}
\bfseries\boldmath configuration & \bfseries\boldmath TPF & \bfseries\boldmath forwards & \bfseries\boldmath tokens & \bfseries\boldmath $\hat R$ ($k/n$) & \bfseries\boldmath $0.05$ & \bfseries\boldmath $0.10$ & \bfseries\boldmath $0.15$ & \bfseries\boldmath $0.20$ \\
\midrule
acc85/semi70 & 4.253 & 17.2 & 73 & 0.178 (118/664) & 1.000 & 1.000 & 0.978 & \underline{\textbf{0.081}} \\
acc85/semi90 & 3.810 & 18.0 & 69 & 0.145 (96/664) & 1.000 & 1.000 & 0.372 & \textbf{1e-4} \\
acc90/semi70 & 3.492 & 19.4 & 68 & 0.139 (92/664) & 1.000 & 0.999 & 0.222 & \textbf{2e-5} \\
acc90/semi90 & 3.349 & 20.6 & 69 & 0.087 (58/664) & 1.000 & 0.153 & \underline{\textbf{9e-7}} & \textbf{1e-15} \\
acc95/semi70 & 2.980 & 23.7 & 71 & 0.081 (54/664) & 1.000 & 0.059 & \textbf{7e-8} & \textbf{3e-17} \\
acc95/semi90 (reference) & 2.664 & 24.2 & 64 & 0.000 (0/664) & \underline{\textbf{2e-15}} & \underline{\textbf{4e-31}} & \textbf{1e-47} & \textbf{4e-65} \\
acc99/semi90 & 2.299 & 32.8 & 75 & 0.018 (12/664) & \textbf{2e-5} & \textbf{2e-17} & \textbf{2e-31} & \textbf{4e-47} \\
acc99/semi70 & 2.225 & 30.7 & 68 & 0.081 (54/664) & 1.000 & 0.059 & \textbf{7e-8} & \textbf{3e-17} \\
\bottomrule
\end{tabular*}
\end{table}

\begin{table}[t]
\caption{Full grid for LLaDA2 math on the reduced prompt set. TPF, forwards and tokens of each request, joint risk $\hat R$ ($k/n$) and exact binomial $p$-values at four budgets. Bold marks the configurations valid under Holm, and underlining the one Redline deploys.}
\label{tab:grid_llada2_math_L256}
\centering\footnotesize
\setlength{\tabcolsep}{2pt}
\begin{tabular*}{\textwidth}{@{\extracolsep{\fill}}lrrrrcccc@{}}
\toprule
 &  &  &  &  & \multicolumn{4}{c}{\bfseries\boldmath $p$-value at budget $\alpha$} \\
\cmidrule(l){6-9}
\bfseries\boldmath configuration & \bfseries\boldmath TPF & \bfseries\boldmath forwards & \bfseries\boldmath tokens & \bfseries\boldmath $\hat R$ ($k/n$) & \bfseries\boldmath $0.05$ & \bfseries\boldmath $0.10$ & \bfseries\boldmath $0.15$ & \bfseries\boldmath $0.20$ \\
\midrule
acc85/semi70 & 5.973 & 131.7 & 787 & 0.057 (29/512) & 0.789 & \underline{\textbf{3e-4}} & \underline{\textbf{3e-11}} & \underline{\textbf{2e-20}} \\
acc85/semi90 & 5.850 & 129.8 & 759 & 0.070 (36/512) & 0.983 & \textbf{0.012} & \textbf{2e-8} & \textbf{2e-16} \\
acc90/semi70 & 5.317 & 146.2 & 778 & 0.064 (33/512) & 0.941 & \textbf{0.003} & \textbf{2e-9} & \textbf{5e-18} \\
acc90/semi90 & 5.245 & 150.9 & 792 & 0.053 (27/512) & 0.659 & \textbf{8e-5} & \textbf{3e-12} & \textbf{1e-21} \\
acc95/semi70 & 4.534 & 165.4 & 750 & 0.051 (26/512) & 0.584 & \textbf{4e-5} & \textbf{1e-12} & \textbf{2e-22} \\
acc95/semi90 (reference) & 4.364 & 159.2 & 695 & 0.000 (0/512) & \underline{\textbf{4e-12}} & \textbf{4e-24} & \textbf{7e-37} & \textbf{2e-50} \\
acc99/semi70 & 3.442 & 207.7 & 715 & 0.049 (25/512) & 0.504 & \textbf{2e-5} & \textbf{3e-13} & \textbf{5e-23} \\
acc99/semi90 & 3.287 & 201.8 & 663 & 0.039 (20/512) & 0.150 & \textbf{2e-7} & \textbf{3e-16} & \textbf{1e-26} \\
\bottomrule
\end{tabular*}
\end{table}

\begin{table}[t]
\caption{Full grid for LLaDA2 code on the reduced prompt set. TPF, forwards and tokens of each request, joint risk $\hat R$ ($k/n$) and exact binomial $p$-values at four budgets. Bold marks the configurations valid under Holm, and underlining the one Redline deploys.}
\label{tab:grid_llada2_code_L256}
\centering\footnotesize
\setlength{\tabcolsep}{2pt}
\begin{tabular*}{\textwidth}{@{\extracolsep{\fill}}lrrrrcccc@{}}
\toprule
 &  &  &  &  & \multicolumn{4}{c}{\bfseries\boldmath $p$-value at budget $\alpha$} \\
\cmidrule(l){6-9}
\bfseries\boldmath configuration & \bfseries\boldmath TPF & \bfseries\boldmath forwards & \bfseries\boldmath tokens & \bfseries\boldmath $\hat R$ ($k/n$) & \bfseries\boldmath $0.05$ & \bfseries\boldmath $0.10$ & \bfseries\boldmath $0.15$ & \bfseries\boldmath $0.20$ \\
\midrule
acc85/semi70 & 7.619 & 28.6 & 218 & 0.143 (60/420) & 1.000 & 0.998 & 0.372 & \underline{\textbf{0.001}} \\
acc85/semi90 & 7.224 & 31.8 & 230 & 0.121 (51/420) & 1.000 & 0.936 & 0.055 & \textbf{1e-5} \\
acc90/semi70 & 6.536 & 32.7 & 214 & 0.117 (49/420) & 1.000 & 0.887 & \underline{\textbf{0.030}} & \textbf{4e-6} \\
acc90/semi90 & 6.339 & 34.2 & 217 & 0.090 (38/420) & 1.000 & 0.290 & \textbf{2e-4} & \textbf{7e-10} \\
acc95/semi70 & 5.492 & 40.1 & 220 & 0.093 (39/420) & 1.000 & 0.349 & \textbf{3e-4} & \textbf{2e-9} \\
acc95/semi90 (reference) & 5.190 & 39.6 & 205 & 0.000 (0/420) & \underline{\textbf{4e-10}} & \underline{\textbf{6e-20}} & \textbf{2e-30} & \textbf{2e-41} \\
acc99/semi70 & 4.182 & 54.7 & 229 & 0.100 (42/420) & 1.000 & 0.541 & \textbf{0.002} & \textbf{2e-8} \\
acc99/semi90 & 3.732 & 53.5 & 200 & 0.043 (18/420) & 0.296 & \textbf{1e-5} & \textbf{1e-12} & \textbf{6e-21} \\
\bottomrule
\end{tabular*}
\end{table}

\subsection{The mean-accuracy rule on the same splits}
\label{app:mean-selector}

We run the mean-accuracy rule on the four main grids under the split protocol of Appendix~\ref{app:validity}, with $K = 1{,}000$ random splits for each grid at calibration fraction $0.5$ and, as checks, $0.3$ and $0.7$. Three selectors see only the calibration half and are scored only on the test half. Redline is the exact binomial test with Holm at $\delta = 0.10$, deploying the fastest valid configuration, or the reference when none is valid. MEAN($t$) deploys the fastest configuration whose calibration-half accuracy is within $t$ points of the reference's calibration-half accuracy, for $t \in \{0, 1, \ldots, 8, 10\}$, and every tolerance is reported. REF always deploys the reference.

Exceedance is the fraction of splits whose deployed configuration has test-half $R > \alpha$. Test-half $R$ is itself an estimate, so a configuration whose population risk sits at the budget crosses it on the test half about half the time, which inflates exceedance for every selector alike. The selectors are therefore compared with each other on the same splits at matched speed. The mean rule is also handed the reference's own calibration accuracy, which is more than a leaderboard reader has.

Table~\ref{tab:mean_selector} prints, for each grid and budget, the mean TPF gain and exceedance of Redline beside the two tolerances that bracket its speed and beside $t = 10$. At matched speed the mean rule exceeds the budget far more often wherever the grid holds configurations near it (SDAR math and LLaDA2 code at $\alpha = 0.10$, SDAR code at $0.15$ and both math grids at $0.05$). The two rules agree where every faster configuration is well under the budget (LLaDA2 math at $\alpha \ge 0.10$, both math grids at $\alpha \ge 0.15$ and LLaDA2 code at $0.15$), and no tolerance reaches the speed of Redline at $\alpha = 0.20$ on either code grid. A calibration fraction of $0.3$ leaves every reading unchanged, and at $0.7$ the SDAR math reading holds ($1.6\%$ against $13.7$ to $37.5\%$ at $\alpha = 0.10$). At both fractions, as at half, no tolerance is at once as fast as Redline on LLaDA2 math and as rarely over the budget on SDAR math at $\alpha = 0.10$.

At every printed budget, each tolerance tested, from $0$ to $10$ points, either gains less TPF than Redline on some grid or exceeds the budget on some grid in at least three times as many of the $1{,}000$ held-out splits and in at least $40$ more of them. At $\alpha = 0.10$, for instance, the tolerances up to four points fall at least $3.4$ points of TPF gain short of Redline on LLaDA2 math, and those from five points up exceed the budget on SDAR math in $71.1$ to $72.4\%$ of held-out splits and in $65.3$ to $100\%$ of splits against the pooled risk, against $3.4\%$ and $0.0\%$ for Redline.

\begin{table}[t]
\caption{Redline against the mean-accuracy rule MEAN($t$) on $1{,}000$ random half splits. Mean TPF gain, exceedance (splits with test-half $R > \alpha$) and deployments of the reference (ref.), all in percent, for the tolerances that bracket the gain of Redline and for $t = 10$.}
\label{tab:mean_selector}
\centering\footnotesize
\setlength{\tabcolsep}{4.5pt}
\begin{tabular*}{\textwidth}{@{\extracolsep{\fill}}c rrr crr crr rr@{}}
\toprule
 & \multicolumn{3}{c}{\textbf{Redline}} & \multicolumn{3}{c}{\begin{tabular}[b]{@{}c@{}}\textbf{MEAN}(\boldmath$t$)\\\textbf{just slower}\end{tabular}} & \multicolumn{3}{c}{\begin{tabular}[b]{@{}c@{}}\textbf{MEAN}(\boldmath$t$)\\\textbf{at or faster}\end{tabular}} & \multicolumn{2}{c}{\textbf{MEAN}(\boldmath$10$)} \\
\cmidrule(lr){2-4}\cmidrule(lr){5-7}\cmidrule(lr){8-10}\cmidrule(lr){11-12}
\boldmath$\alpha$ & \textbf{gain} & \textbf{exc.} & \textbf{ref.} & \boldmath$t$ & \textbf{gain} & \textbf{exc.} & \boldmath$t$ & \textbf{gain} & \textbf{exc.} & \textbf{gain} & \textbf{exc.} \\
\midrule
\rowcolor{groupyellow}\multicolumn{12}{@{}c@{}}{\textit{LLaDA2 math}\strut} \\
0.05 & 0.0 & 0.1 & 99.9 & \multicolumn{3}{c}{none slower} & 0 & 0.6 & 5.7 & 37.5 & 99.6 \\
0.10 & 36.6 & 0.1 & 0.0 & 5 & 36.6 & 0.1 & 6 & 37.3 & 0.1 & 37.5 & 0.1 \\
0.15 & 37.5 & 0.0 & 0.0 & 6 & 37.3 & 0.0 & 7 & 37.5 & 0.0 & 37.5 & 0.0 \\
0.20 & 37.5 & 0.0 & 0.0 & 6 & 37.3 & 0.0 & 7 & 37.5 & 0.0 & 37.5 & 0.0 \\
\rowcolor{groupyellow}\multicolumn{12}{@{}c@{}}{\textit{LLaDA2 code}\strut} \\
0.05 & 0.0 & 0.0 & 100.0 & \multicolumn{3}{c}{none slower} & 0 & 0.0 & 0.1 & 33.6 & 99.9 \\
0.10 & 1.5 & 1.3 & 91.1 & 2 & 1.0 & 2.9 & 3 & 3.7 & 10.2 & 33.6 & 73.3 \\
0.15 & 24.7 & 0.4 & 0.1 & 7 & 23.9 & 0.5 & 8 & 29.1 & 1.2 & 33.6 & 3.6 \\
0.20 & 40.2 & 0.1 & 0.0 & 10 & 33.6 & 0.1 & \multicolumn{3}{c}{none as fast} & 33.6 & 0.1 \\
\rowcolor{groupyellow}\multicolumn{12}{@{}c@{}}{\textit{SDAR math}\strut} \\
0.05 & 0.0 & 0.0 & 100.0 & \multicolumn{3}{c}{none slower} & 0 & 1.0 & 7.0 & 36.6 & 100.0 \\
0.10 & 8.3 & 3.4 & 39.5 & 1 & 4.5 & 6.4 & 2 & 12.0 & 21.1 & 36.6 & 71.1 \\
0.15 & 36.5 & 0.0 & 0.0 & 7 & 36.3 & 0.0 & 8 & 36.5 & 0.0 & 36.6 & 0.0 \\
0.20 & 36.6 & 0.0 & 0.0 & 8 & 36.5 & 0.0 & 10 & 36.6 & 0.0 & 36.6 & 0.0 \\
\rowcolor{groupyellow}\multicolumn{12}{@{}c@{}}{\textit{SDAR code}\strut} \\
0.05 & 0.0 & 0.0 & 100.0 & \multicolumn{3}{c}{none slower} & 0 & 0.0 & 0.0 & 36.5 & 100.0 \\
0.10 & 1.3 & 5.8 & 92.2 & 2 & 0.3 & 1.2 & 3 & 1.6 & 4.9 & 36.5 & 70.3 \\
0.15 & 25.8 & 4.1 & 0.0 & 7 & 25.6 & 7.3 & 8 & 28.3 & 18.3 & 36.5 & 36.6 \\
0.20 & 46.9 & 6.8 & 0.0 & 10 & 36.5 & 1.3 & \multicolumn{3}{c}{none as fast} & 36.5 & 1.3 \\
\bottomrule
\end{tabular*}
\end{table}

\textbf{Stronger baselines on the same splits.} Table~\ref{tab:baselines} runs every selector a reader might prefer on the same $1{,}000$ half splits and reports exceedance under both measures, the test-half risk of the deployed configuration and its risk pooled over all $n$ prompts, which the calibration half also informs and which a configuration selected on that half therefore reads optimistically. For each grid and budget, it gives the mean TPF gain of the deployed configuration, counting deployments of the reference as zero, and both exceedances. Its columns are Redline, Bonferroni (BONF), the fixed-sequence walk (FIXSEQ), the uncorrected test (UNCORR), the plug-in rule (PLUGIN), the net-drop test NETHOLM with a Hoeffding (H), Hoeffding--Bentkus (HB) or betting (WSR) $p$-value, the mean rule with a normal bound on the drop (MEANCI), and MEAN($t$) at the two tolerances that bracket the speed of Redline in Table~\ref{tab:mean_selector} and at $t = 10$. Where no tolerance from $0$ to $10$ runs slower than Redline, or none runs as fast, the bracket does not exist and the table says so.

Bonferroni and the fixed-sequence walk in the order of the design thresholds, the two other corrections with a family-wise guarantee, exceed the budget in at most $1.6\%$ of splits against the pooled risk, and Bonferroni deploys a slower configuration than Redline wherever the two differ. Dropping the multiplicity correction, the uncorrected exact test at $\delta$ for each configuration, raises held-out exceedance to $17.5\%$ on SDAR math at $\alpha = 0.10$ and $18.7\%$ on SDAR code at $0.15$. Dropping the margin as well, the plug-in rule $\hat R \le \alpha$ on the calibration half exceeds the budget in $28.5$ to $58.4\%$ of splits pooled and $28.9$ to $64.1\%$ held out on LLaDA2 math at $0.05$, LLaDA2 code at $0.10$ and $0.15$ and SDAR math at $0.10$.

Two selectors that bound the net accuracy drop rather than the joint risk, Holm at $\delta$ with the betting $p$-value of \citet{waudbysmith2024betting} on the rescaled drop and the mean rule with a one-sided $90\%$ normal bound on the drop, keep their own quantity within budget in all but at most $2.5\%$ of splits and still exceed the joint budget in $91.4\%$ and $99.1\%$ of splits on SDAR math at $0.10$, the gap between the two risks that our guarantee closes. The Hoeffding and Hoeffding--Bentkus variants \citep{bates2021rcps} of that test almost always deploy the reference, since those bounds are loose for a three-valued loss.

\begin{table}[p]
\caption{Every selector on the same $1{,}000$ random half splits ($\delta = 0.10$). Mean TPF gain of the deployed configuration, and exceedance on the test half and against the pooled risk, in percent.}
\label{tab:baselines}
\centering\scriptsize\setlength{\tabcolsep}{1.4pt}
\begin{tabular*}{\textwidth}{@{\extracolsep{\fill}}cl rrrrr rrr r rrr@{}}
\toprule
 & & & & & & & \multicolumn{3}{c}{\textbf{NETHOLM}} & & \multicolumn{3}{c}{\textbf{MEAN}(\boldmath$t$)} \\
\cmidrule(lr){8-10}\cmidrule(l){12-14}
\boldmath$\alpha$ & \textbf{measure} & \textbf{Redline} & \textbf{BONF} & \textbf{FIXSEQ} & \textbf{UNCORR} & \textbf{PLUGIN} & \textbf{H} & \textbf{HB} & \textbf{WSR} & \textbf{MEANCI} & \textbf{slower} & \textbf{faster} & \boldmath$t = 10$ \\
\midrule
\grouprow{14}{LLaDA2 math ($n = 1{,}012$)} \\
0.05 & gain & 0.0 & 0.0 & 0.1 & 0.2 & 7.8 & 0.0 & 0.0 & 8.6 & 27.0 & \multirow{3}{*}{\textit{none}} & 0.6 & 37.5 \\
 & test & 0.1 & 0.1 & 1.6 & 2.4 & 58.4 & 0.0 & 0.0 & 47.9 & 89.6 &  & 5.7 & 99.6 \\
 & pool & 0.1 & 0.1 & 1.6 & 2.4 & 58.4 & 0.0 & 0.0 & 50.3 & 99.5 &  & 5.7 & 100.0 \\
0.10 & gain & 36.6 & 32.7 & 36.7 & 37.0 & 37.5 & 0.0 & 0.0 & 37.5 & 37.5 & 36.6 & 37.3 & 37.5 \\
 & test & 0.1 & 0.1 & 0.1 & 0.1 & 0.1 & 0.0 & 0.0 & 0.1 & 0.1 & 0.1 & 0.1 & 0.1 \\
 & pool & 0.0 & 0.0 & 0.0 & 0.0 & 0.0 & 0.0 & 0.0 & 0.0 & 0.0 & 0.0 & 0.0 & 0.0 \\
0.15 & gain & 37.5 & 37.5 & 37.5 & 37.5 & 37.5 & 29.1 & 37.3 & 37.5 & 37.5 & 37.3 & 37.5 & 37.5 \\
 & test & 0.0 & 0.0 & 0.0 & 0.0 & 0.0 & 0.0 & 0.0 & 0.0 & 0.0 & 0.0 & 0.0 & 0.0 \\
 & pool & 0.0 & 0.0 & 0.0 & 0.0 & 0.0 & 0.0 & 0.0 & 0.0 & 0.0 & 0.0 & 0.0 & 0.0 \\
0.20 & gain & 37.5 & 37.5 & 37.5 & 37.5 & 37.5 & 37.5 & 37.5 & 37.5 & 37.5 & 37.3 & 37.5 & 37.5 \\
 & test & 0.0 & 0.0 & 0.0 & 0.0 & 0.0 & 0.0 & 0.0 & 0.0 & 0.0 & 0.0 & 0.0 & 0.0 \\
 & pool & 0.0 & 0.0 & 0.0 & 0.0 & 0.0 & 0.0 & 0.0 & 0.0 & 0.0 & 0.0 & 0.0 & 0.0 \\
\grouprow{14}{LLaDA2 code ($n = 542$)} \\
0.05 & gain & 0.0 & 0.0 & 0.0 & 0.0 & 0.2 & 0.0 & 0.0 & 0.1 & 1.8 & \multirow{3}{*}{\textit{none}} & 0.0 & 33.6 \\
 & test & 0.0 & 0.0 & 0.0 & 0.0 & 1.0 & 0.0 & 0.0 & 1.0 & 12.1 &  & 0.1 & 99.9 \\
 & pool & 0.0 & 0.0 & 0.0 & 0.0 & 1.0 & 0.0 & 0.0 & 1.0 & 12.1 &  & 0.1 & 100.0 \\
0.10 & gain & 1.5 & 1.5 & 0.4 & 8.2 & 21.5 & 0.0 & 0.0 & 9.4 & 25.5 & 1.0 & 3.7 & 33.6 \\
 & test & 1.3 & 1.3 & 1.0 & 2.2 & 28.9 & 0.0 & 0.0 & 11.9 & 52.2 & 2.9 & 10.2 & 73.3 \\
 & pool & 0.0 & 0.0 & 0.0 & 0.4 & 28.5 & 0.0 & 0.0 & 5.3 & 52.4 & 0.0 & 0.7 & 97.8 \\
0.15 & gain & 24.7 & 21.5 & 17.7 & 30.4 & 36.3 & 0.0 & 0.0 & 32.7 & 35.7 & 23.9 & 29.1 & 33.6 \\
 & test & 0.4 & 0.2 & 1.0 & 1.1 & 30.6 & 0.0 & 0.0 & 9.5 & 23.7 & 0.5 & 1.2 & 3.6 \\
 & pool & 0.2 & 0.0 & 0.2 & 0.3 & 30.6 & 0.0 & 0.0 & 9.6 & 23.7 & 0.0 & 0.0 & 3.6 \\
0.20 & gain & 40.2 & 35.8 & 40.0 & 40.4 & 42.3 & 2.6 & 12.0 & 41.3 & 42.2 & 33.6 & \multirow{3}{*}{\textit{none}} & 33.6 \\
 & test & 0.1 & 0.1 & 0.1 & 0.1 & 0.1 & 0.0 & 0.0 & 0.1 & 0.1 & 0.1 &  & 0.1 \\
 & pool & 0.0 & 0.0 & 0.0 & 0.0 & 0.0 & 0.0 & 0.0 & 0.0 & 0.0 & 0.0 &  & 0.0 \\
\grouprow{14}{SDAR math ($n = 1{,}012$)} \\
0.05 & gain & 0.0 & 0.0 & 0.0 & 0.0 & 0.1 & 0.0 & 0.0 & 2.5 & 18.2 & \multirow{3}{*}{\textit{none}} & 1.0 & 36.6 \\
 & test & 0.0 & 0.0 & 0.0 & 0.0 & 0.8 & 0.0 & 0.0 & 16.7 & 88.9 &  & 7.0 & 100.0 \\
 & pool & 0.0 & 0.0 & 0.0 & 0.0 & 0.8 & 0.0 & 0.0 & 16.7 & 88.9 &  & 7.0 & 100.0 \\
0.10 & gain & 8.3 & 6.7 & 6.2 & 15.1 & 29.2 & 0.0 & 0.1 & 35.7 & 36.5 & 4.5 & 12.0 & 36.6 \\
 & test & 3.4 & 2.4 & 2.8 & 17.5 & 64.1 & 0.0 & 0.3 & 71.6 & 71.4 & 6.4 & 21.1 & 71.1 \\
 & pool & 0.0 & 0.0 & 0.0 & 1.2 & 30.4 & 0.0 & 0.0 & 91.4 & 99.1 & 0.2 & 2.6 & 100.0 \\
0.15 & gain & 36.5 & 35.2 & 36.5 & 36.5 & 36.6 & 18.8 & 34.0 & 36.6 & 36.6 & 36.3 & 36.5 & 36.6 \\
 & test & 0.0 & 0.0 & 0.0 & 0.0 & 0.0 & 0.0 & 0.0 & 0.0 & 0.0 & 0.0 & 0.0 & 0.0 \\
 & pool & 0.0 & 0.0 & 0.0 & 0.0 & 0.0 & 0.0 & 0.0 & 0.0 & 0.0 & 0.0 & 0.0 & 0.0 \\
0.20 & gain & 36.6 & 36.6 & 36.6 & 36.6 & 36.6 & 36.6 & 36.6 & 36.6 & 36.6 & 36.5 & 36.6 & 36.6 \\
 & test & 0.0 & 0.0 & 0.0 & 0.0 & 0.0 & 0.0 & 0.0 & 0.0 & 0.0 & 0.0 & 0.0 & 0.0 \\
 & pool & 0.0 & 0.0 & 0.0 & 0.0 & 0.0 & 0.0 & 0.0 & 0.0 & 0.0 & 0.0 & 0.0 & 0.0 \\
\grouprow{14}{SDAR code ($n = 664$)} \\
0.05 & gain & 0.0 & 0.0 & 0.0 & 0.0 & 0.0 & 0.0 & 0.0 & 0.0 & 1.2 & \multirow{3}{*}{\textit{none}} & 0.0 & 36.5 \\
 & test & 0.0 & 0.0 & 0.0 & 0.0 & 0.0 & 0.0 & 0.0 & 0.1 & 5.0 &  & 0.0 & 100.0 \\
 & pool & 0.0 & 0.0 & 0.0 & 0.0 & 0.0 & 0.0 & 0.0 & 0.1 & 5.0 &  & 0.0 & 100.0 \\
0.10 & gain & 1.3 & 1.2 & 5.1 & 7.1 & 23.9 & 0.0 & 0.0 & 12.2 & 26.2 & 0.3 & 1.6 & 36.5 \\
 & test & 5.8 & 5.8 & 8.4 & 13.6 & 10.9 & 0.0 & 0.0 & 12.3 & 14.7 & 1.2 & 4.9 & 70.3 \\
 & pool & 0.0 & 0.0 & 0.0 & 0.0 & 0.3 & 0.0 & 0.0 & 1.1 & 7.5 & 0.0 & 0.0 & 70.1 \\
0.15 & gain & 25.8 & 25.3 & 27.3 & 27.7 & 38.0 & 0.0 & 0.1 & 34.5 & 43.8 & 25.6 & 28.3 & 36.5 \\
 & test & 4.1 & 1.5 & 16.2 & 18.7 & 37.4 & 0.0 & 0.0 & 34.5 & 36.5 & 7.3 & 18.3 & 36.6 \\
 & pool & 0.0 & 0.0 & 0.0 & 0.0 & 2.5 & 0.0 & 0.0 & 7.2 & 11.8 & 0.1 & 0.1 & 1.4 \\
0.20 & gain & 46.9 & 38.0 & 47.7 & 47.8 & 58.5 & 11.4 & 23.9 & 56.0 & 58.8 & 36.5 & \multirow{3}{*}{\textit{none}} & 36.5 \\
 & test & 6.8 & 2.5 & 6.8 & 6.8 & 6.8 & 0.1 & 0.6 & 6.8 & 6.8 & 1.3 &  & 1.3 \\
 & pool & 0.0 & 0.0 & 0.0 & 0.0 & 0.0 & 0.0 & 0.0 & 0.0 & 0.0 & 0.0 &  & 0.0 \\
\bottomrule
\end{tabular*}
\end{table}

\subsection{Held-out re-measurement of the deployed configurations}
\label{app:heldout-deploy}

Table~\ref{tab:unlock} reports the TPF gain and net accuracy change of each deployed configuration on the calibration prompts that selected it. Under the split protocol of Appendix~\ref{app:validity}, with $K = 1{,}000$ half splits, Redline runs on one half and the deployed configuration's TPF gain, net accuracy change and joint risk are measured on the other half. The tokens and forwards of each prompt come from the decode trajectories. Table~\ref{tab:heldout} prints, for each grid and budget, the full-sample deployment and its in-sample numbers beside the held-out mean and the $2.5$ to $97.5$ percentile range over the splits that deploy that same configuration, together with the number of splits that deploy the reference.

At $\alpha = 0.10$ the LLaDA2 math configuration is deployed in $894$ of $1{,}000$ splits and reads $+37.6\%$ TPF $[+34.1, +41.3]$ and $-4.2$pp $[-5.9, -2.6]$ held out, against $+37.5\%$ and $-4.0$pp in sample. The SDAR math configuration is deployed in $548$ splits and reads $+13.4\%$ $[+7.6, +19.7]$ and $-3.1$pp $[-4.7, -1.5]$ against $+13.1\%$ and $-2.5$pp. Every in-sample TPF gain lies inside its held-out range.

\begin{table}[t]
\caption{Held-out re-measurement over $1{,}000$ random half splits of the configuration that Redline deploys on one half, measured on the other. Splits deploying the reference and the full-sample configuration, and held-out mean with $95\%$ percentile range, gain in percent and net accuracy in points.}
\label{tab:heldout}
\centering\footnotesize
\setlength{\tabcolsep}{2pt}
\begin{tabular*}{\textwidth}{@{\extracolsep{\fill}}cl rr rr cc@{}}
\toprule
 &  & \multicolumn{2}{c}{\bfseries\boldmath in sample} & \multicolumn{2}{c}{\bfseries\boldmath splits} & \multicolumn{2}{c}{\bfseries\boldmath held out} \\
\cmidrule(lr){3-4}\cmidrule(lr){5-6}\cmidrule(lr){7-8}
\bfseries\boldmath $\alpha$ & \bfseries\boldmath deployed & \bfseries\boldmath gain & \bfseries\boldmath net & \bfseries\boldmath ref. & \bfseries\boldmath same & \bfseries\boldmath gain & \bfseries\boldmath net \\
\midrule
\rowcolor{groupyellow}\multicolumn{8}{@{}c@{}}{\textit{LLaDA2 math}\strut} \\
0.05 & reference & +0.0 & 0.0 & 999 & 999 & +0.0 & 0.0 \\
0.10 & acc85/semi70 & +37.5 & $-$4.0 & 0 & 894 & +37.6 [+34.1, +41.3] & $-$4.2 [$-$5.9, $-$2.6] \\
0.15 & acc85/semi70 & +37.5 & $-$4.0 & 0 & 1000 & +37.6 [+34.1, +41.4] & $-$4.0 [$-$5.9, $-$2.2] \\
0.20 & acc85/semi70 & +37.5 & $-$4.0 & 0 & 1000 & +37.6 [+34.1, +41.4] & $-$4.0 [$-$5.9, $-$2.2] \\
\rowcolor{groupyellow}\multicolumn{8}{@{}c@{}}{\textit{LLaDA2 code}\strut} \\
0.05 & reference & +0.0 & 0.0 & 1000 & 1000 & +0.0 & 0.0 \\
0.10 & reference & +0.0 & 0.0 & 911 & 911 & +0.0 & 0.0 \\
0.15 & acc85/semi90 & +33.7 & $-$7.0 & 1 & 465 & +33.8 [+26.6, +41.3] & $-$8.2 [$-$10.3, $-$5.9] \\
0.20 & acc85/semi70 & +42.4 & $-$13.1 & 0 & 755 & +42.7 [+33.2, +52.9] & $-$13.8 [$-$16.6, $-$11.4] \\
\rowcolor{groupyellow}\multicolumn{8}{@{}c@{}}{\textit{SDAR math}\strut} \\
0.05 & reference & +0.0 & 0.0 & 1000 & 1000 & +0.0 & 0.0 \\
0.10 & acc90/semi90 & +13.1 & $-$2.5 & 395 & 548 & +13.4 [+7.6, +19.7] & $-$3.1 [$-$4.7, $-$1.5] \\
0.15 & acc85/semi70 & +36.6 & $-$4.6 & 0 & 994 & +36.6 [+29.1, +44.5] & $-$4.7 [$-$7.3, $-$2.2] \\
0.20 & acc85/semi70 & +36.6 & $-$4.6 & 0 & 1000 & +36.6 [+29.1, +44.5] & $-$4.7 [$-$7.3, $-$2.2] \\
\rowcolor{groupyellow}\multicolumn{8}{@{}c@{}}{\textit{SDAR code}\strut} \\
0.05 & reference & +0.0 & 0.0 & 1000 & 1000 & +0.0 & 0.0 \\
0.10 & reference & +0.0 & 0.0 & 922 & 922 & +0.0 & 0.0 \\
0.15 & acc90/semi90 & +25.7 & $-$5.0 & 0 & 938 & +25.5 [+17.4, +33.8] & $-$5.0 [$-$7.5, $-$2.4] \\
0.20 & acc85/semi70 & +59.7 & $-$14.0 & 0 & 311 & +58.8 [+37.8, +79.8] & $-$15.8 [$-$18.1, $-$13.9] \\
\bottomrule
\end{tabular*}
\end{table}

\subsection{A LLaDA2 math grid of 50 configurations}
\label{app:grids-biggrid}

The LLaDA2 math grid of Table~\ref{tab:unlock} tests $m = 7$ configurations, Holm's correction is cheap at that size, and from $\alpha = 0.10$ on its deployed configuration is the most aggressive one of the grid, so the grid itself bounds the attainable gain. To measure what a larger grid costs and returns, we served a grid of $50$ configurations over the engine's three lossy thresholds, accept $\{0.75, 0.80, 0.85, 0.90, 0.95\}$ $\times$ semi-completion $\{0.5, 0.6, 0.7, 0.8, 0.9\}$ $\times$ add-block $\{0.10, 0.30\}$, on the same math set, scorer and serving batch ($n = 1{,}012$). Every configuration, the reference acc95/semi90/add0.10 included, was served anew for this grid, so every risk and gain below is relative to that reference. The add $0.30$ configurations run within $5\%$ TPF of their add $0.10$ partners at every accept and semi-completion setting (ratios $0.973$ to $1.023$), so the second add level doubles the grid with near-duplicate hypotheses, which stresses the multiplicity correction.

\begin{table}[t]
\caption{The LLaDA2 math grid of $50$ configurations tested by Redline as three nested grids of the same served outputs ($n = 1{,}012$, $\delta = 0.10$). For each grid and budget, the deployed configuration, the number of valid configurations, its violations $k$, joint risk, TPF, gain and net accuracy change.}
\label{tab:biggrid}
\centering\small
\begin{tabular*}{\textwidth}{@{\extracolsep{\fill}}cl cccc cc@{}}
\toprule
\bfseries\boldmath $\alpha$ & \bfseries\boldmath deployed & \bfseries\boldmath valid & \bfseries\boldmath $k$ & \bfseries\boldmath $\hat R$ & \bfseries\boldmath TPF & \bfseries\boldmath TPF gain (\%) & \bfseries\boldmath net accuracy (pp) \\
\midrule
\grouprow{8}{The six configurations shared with the main grid, $m = 6$} \\
0.05 & reference & 1 of 6 & 0 & 0.000 & 4.42 & $+0.0$ & $+0.0$ \\
0.10 & acc85/semi70/add0.10 & 6 of 6 & 53 & 0.052 & 5.90 & $+33.4$ & $-1.5$ \\
0.15 & acc85/semi70/add0.10 & 6 of 6 & 53 & 0.052 & 5.90 & $+33.4$ & $-1.5$ \\
0.20 & acc85/semi70/add0.10 & 6 of 6 & 53 & 0.052 & 5.90 & $+33.4$ & $-1.5$ \\
\grouprow{8}{The add $0.10$ slice, $m = 25$} \\
0.05 & reference & 1 of 25 & 0 & 0.000 & 4.42 & $+0.0$ & $+0.0$ \\
0.10 & acc80/semi50/add0.10 & 19 of 25 & 79 & 0.078 & 6.84 & $+54.7$ & $-4.0$ \\
0.15 & acc75/semi50/add0.10 & 25 of 25 & 99 & 0.098 & 7.60 & $+72.0$ & $-5.9$ \\
0.20 & acc75/semi50/add0.10 & 25 of 25 & 99 & 0.098 & 7.60 & $+72.0$ & $-5.9$ \\
\grouprow{8}{The whole grid, $m = 50$} \\
0.05 & reference & 1 of 50 & 0 & 0.000 & 4.42 & $+0.0$ & $+0.0$ \\
0.10 & acc75/semi80/add0.30 & 36 of 50 & 74 & 0.073 & 7.31 & $+65.4$ & $-3.4$ \\
0.15 & acc75/semi50/add0.10 & 50 of 50 & 99 & 0.098 & 7.60 & $+72.0$ & $-5.9$ \\
0.20 & acc75/semi50/add0.10 & 50 of 50 & 99 & 0.098 & 7.60 & $+72.0$ & $-5.9$ \\
\bottomrule
\end{tabular*}
\end{table}

Table~\ref{tab:biggrid} runs Redline at $\delta = 0.10$ on three nested grids of the same served outputs, the six configurations the grid shares with the main grid (accept $\{0.85, 0.90, 0.95\}$ $\times$ semi-completion $\{0.7, 0.9\}$ at add $0.10$, $m = 6$), the add $0.10$ slice ($m = 25$) and the whole grid ($m = 50$). The grid of six deploys the same configuration as the main grid, acc85/semi70, at every budget from $0.10$ on, at $+33.4\%$ against the reference of this grid.

On the whole grid the configuration deployed at $\alpha = 0.10$ moves to acc75/semi80/add0.30 at $+65.4\%$ ($k = 74$, $36$ of $50$ configurations valid), and at $\alpha = 0.15$ and $0.20$ to acc75/semi50/add0.10 at $+72.0\%$ with all $50$ valid, with net accuracy changes of $-3.4$ and $-5.9$pp beside the risk bound as in Table~\ref{tab:unlock}. On the add $0.10$ slice alone the configuration deployed at $\alpha = 0.10$ is acc80/semi50 at $+54.7\%$ ($19$ of $25$ valid), so at this budget the step from $25$ to $50$ hypotheses added a faster valid configuration rather than removing one.

\textbf{The price of the correction.} Holm and Bonferroni deploy the same configuration at every printed budget except $\alpha = 0.10$ in the grid of $25$. At $\alpha = 0.10$ on the whole grid the deployed configuration's $p = 0.0018$ sits under Bonferroni's $\delta/50 = 0.0020$ and under Holm's threshold at its rank, $0.0056$. On the fine grid of $30$ budgets, the configuration deployed by Holm is faster than Bonferroni's at three budgets in the grid of $50$ ($\alpha = 0.09$, $0.11$ and $0.12$, by $9.1$, $3.0$ and $3.1$ points of TPF gain), at three in the grid of $25$ ($0.10$ to $0.12$, at most $6.1$ points) and at none in the grid of six.

Against an uncorrected test of each configuration at $p \le \delta$, which carries no family-wise guarantee, Holm gives up $3.0$ points at $\alpha = 0.10$ on the whole grid (that test would deploy acc75/semi70/add0.10 at $+68.4\%$) and nothing at $0.15$ or $0.20$. A grid seven times larger therefore costs at most $3.0$ points of gain at the printed budgets and reaches configurations the small grid does not contain.

\textbf{Held out.} Under the split protocol of Appendix~\ref{app:heldout-deploy} on the grid of $50$, the deployed configuration's test-half risk exceeds the budget in $5.3\%$ of splits at $\alpha = 0.10$ and in no split at $0.15$ and $0.20$, where every split deploys the same configuration and reads $+72.0\%$ $[+67.9, +76.3]$ held out.

\textbf{Grid size.} For Figure~\ref{fig:gridsize} we test $1{,}000$ random sub-grids of $8$, $16$ and $32$ configurations of the grid of $50$, each holding the grid's reference and $m - 1$ of its other configurations, with Redline at $\delta = 0.10$ on all $n$ prompts, and place the designed nested grids of Table~\ref{tab:biggrid} beside them. At $\alpha = 0.10$, $0.15$ and $0.20$ no sub-grid deploys the reference. At $\alpha = 0.10$ the mean deployed gain rises from $+53.3\%$ at $m = 8$ through $+57.0\%$ and $+61.3\%$ to the whole grid's $+65.4\%$, and at $\alpha = 0.15$ and $0.20$, whose deployed configurations coincide in every sub-grid, from $+65.4\%$ to $+72.0\%$. Each step is larger than its Monte Carlo standard error, which is at most $0.32$ points. The designed grid of six reaches $+33.4\%$ at every budget from $0.10$, so a larger grid adds valid configurations faster than the correction removes them.

\section{Speculative Decoding and Quantization}
\label{app:generality}

\subsection{Typical acceptance}
\label{app:gen-typical}

The lossy acceptance rule is the typical acceptance of Medusa \citep{cai2024medusa}, which accepts a draft token $x$ when
\begin{equation}
p_{\text{target}}(x) \;>\; \min\bigl(\epsilon,\; a \cdot \exp(-H(p_{\text{target}}))\bigr),
\end{equation}
with $H$ the entropy of the target's next-token distribution and the entropy coefficient coupled as $a = \sqrt{\epsilon}$. A lower $\epsilon$ accepts more draft tokens, which is faster and lossier, so $\epsilon$ is the setting that Redline searches, not a risk level. We apply the published rule on top of greedy verification rather than inside a deployed serving stack. The tested grid is the lossless reference and the six lossy settings $\epsilon \in \{0.4, 0.3, 0.2, 0.1, 0.05, 0.02\}$. Five further settings, $\epsilon \in \{0.8, 0.7, 0.6, 0.5, 0.45\}$, fill in the frontier on the same $512$ calibration prompts. Testing all eleven with Holm is strictly more conservative than testing the six, and it gives identical deployments and gains at $\alpha \in \{0.05, 0.10, 0.15, 0.20\}$ on both draft pairs, for instance $\epsilon = 0.1$ at $+13.8\%$ on the Llama pair at $\alpha = 0.10$.

\subsection{The lossless reference}
\label{app:gen-lossless}

The reference accepts a draft token only when it equals the target's argmax, which reproduces the target's greedy output in exact arithmetic. Its reference-relative risk is therefore zero by construction, and its accepted tokens for each target forward are the speedup that costs nothing ($4.14$ on the Llama pair, $+314\%$ over autoregressive decoding). Standard speculative decoding is lossless in the same sense, since rejection sampling reproduces the target distribution, so testing it would be vacuous. The guarantee is informative only for a lossy acceptance rule, which trades fidelity for speed.

\begin{figure}[t]
\centering
\includegraphics[width=\linewidth]{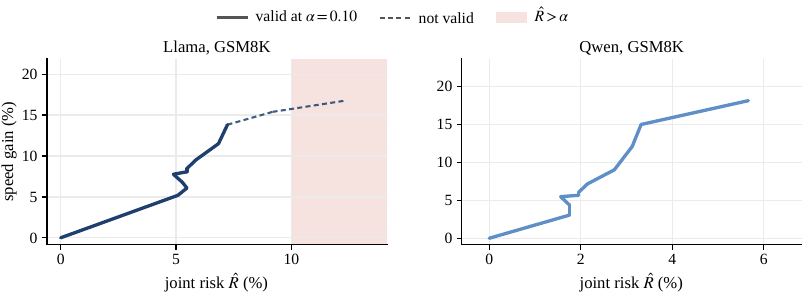}
\caption{Typical acceptance for two draft pairs on GSM8K. Each vertex is a threshold setting at its empirical joint risk and gain in accepted tokens for each target forward over lossless verification. Solid segments join settings valid at $\alpha = 0.10$ and end at the one Redline deploys.}
\label{fig:pareto}
\end{figure}

With each grid at its own $\delta$, the Llama setting deployed on GSM8K at $\alpha = 0.10$ gains $+13.8\%$ accepted tokens for each target forward over lossless verification (net accuracy $-3.3$pp), and Qwen on GSM8K gains from $\alpha = 0.04$ on, with $+15.0\%$ at $\alpha = 0.05$ (net $+0.4$pp) and $+18.1\%$ at $\alpha = 0.10$. At $\alpha = 0.10$ the deployed settings are $\epsilon = 0.1$ for Llama and $\epsilon = 0.02$ for Qwen. Table~\ref{tab:unlock} lists the deployments of both methods at $\alpha = 0.10$, $0.15$ and $0.20$, and Figure~\ref{fig:pareto} places every setting by its risk and its gain.

\subsection{Joint risk against net accuracy}
\label{app:gen-gain}

The joint risk bounds the net accuracy drop from above. The net drop equals $R - G$ with $G = \Pr(\text{reference wrong} \wedge \lambda\ \text{right}) \ge 0$, so $R \le \alpha$ bounds the net drop of every valid configuration. The two pairs split cleanly. All eleven lossy Llama settings lose net accuracy ($-1.6$ to $-9.0$pp), while ten of the eleven Qwen settings raise it and only $\epsilon = 0.02$ loses. The Qwen setting deployed at $\alpha = 0.05$ has $\hat R = 0.033$ and a net change of $+0.4$pp, because its gain term $G = 0.037$ exceeds its risk. Draft agreement does not explain the split. Token-level agreement between the draft and target argmax, measured in lossless mode with four draft tokens, is higher for Llama ($90.5\%$, $4.14$ accepted tokens for each target forward) than for Qwen ($88.7\%$, $3.99$), yet Llama carries the higher risk. The operating point that passes depends instead on whether a divergent token changes the final answer, which is exactly what the risk measures.

\subsection{Quantization}
\label{app:gen-quant}

The quantization grid holds bf16, int8, nf4 and fp4 weights for Llama-3.1-8B on GSM8K ($n = 512$), with bf16 as the reference. At $\alpha = 0.10$ all three quantized configurations are valid (int8 at $\hat R = 0.045$ and net $+0.2$pp), and the deployment rule of this grid, the smallest memory among valid configurations, selects nf4. The nf4 configuration ties with fp4 in theoretical size, is listed before it and has the lower empirical risk of the two ($\hat R = 0.064$ against $0.078$). Measured during cached generation on $20$ prompts, nf4 reduces weight memory by $2.81\times$ and peak memory by $2.75\times$, short of the theoretical $4\times$ because the embeddings and the output head stay in bf16.

\subsection{Answer extraction}
\label{app:gen-scorer}

The math scorer reads the instructed answer marker and falls back to the last number of the output. On the Llama pair's generations for the full GSM8K test set ($1{,}319$ problems), a strict scorer that accepts only the instructed marker, allowing a currency sign and thousands separators, reads within $0.3$pp of this scorer on the lossless reference and within $0.08$pp on the deployed configuration.

\section{Engine and Measurement Details}
\label{app:engine}

The serving engine \citep{jin2026mbd} keeps a fixed number of block slots for each request. A slot holds a placeholder (mask-filled and not yet admitted), an active block (admitted and being denoised), a committed block awaiting its cache write, or a cached block. The admission threshold $\tau_{\text{add}}$ turns a placeholder into an active block, the accept and semi-completion thresholds decide which positions an active block resolves in a step, and the commit sweep of Appendix~\ref{app:mec} commits resolved blocks in order. Every forward runs over the whole window of slots, placeholders included.

\subsection{Occupancy}
\label{app:engine-occupancy}

The occupancy figures of Section~\ref{sec:motivation-space} separate three step-level quantities on math decode traces (GSM8K and MATH-500).
\textbf{Mask-holding slots} are the slots of the forward window that contain any mask token, placeholders included. A four-slot buffer reads about $3.9$ and an eight-slot buffer about $7.8$.

\textbf{Admitted blocks} are the active blocks together with committed blocks awaiting their cache write, which are the blocks in flight. They number about $2.0$ at buffers $4$ and $8$ in both families, and about $1.8$ at buffer $2$, where the buffer itself caps them.

\textbf{Partially resolved blocks} have between $1$ and $31$ of their $32$ positions masked. They number about $1.1$ everywhere, $1.14$ on the SDAR trace of the reference configuration and $1.13$ on the LLaDA2 trace, with at most two in over $99\%$ of steps. Across buffers $2$, $4$ and $8$ the LLaDA2 mean reads $1.10$, $1.13$ and $1.13$, and the SDAR mean reads $1.11$ at buffer $2$ against $1.16$ at buffer $4$.

The buffer sweeps use $128$ prompts in each benchmark on the same engine build for both families. Snapshots record the state after the commit sweep, and steps that encode the prompt are excluded throughout.

\subsection{No-op accounting}
\label{app:engine-noop}

The no-op share of Section~\ref{sec:motivation-space} is a fraction of active block-forwards, the forwards of admitted blocks that are not yet committed. Over $512$ requests in each family it is $37.3\%$ of $140{,}812$ active block-forwards on LLaDA2 and $38.3\%$ of $120{,}287$ on SDAR. Together with the $19{,}906$ one-step cache writes of committed blocks, these $261{,}099$ active block-forwards make up the $281{,}005$ admitted block-steps at which Appendix~\ref{app:mec-validation} checks the commit invariant. Placeholder slots are forwarded by the static window and enter neither count, so the share understates the total waste of the window.

\subsection{Engine configuration}
\label{app:engine-config}

Every configuration in our grids is a real serving run, and each run archives its fully resolved configuration beside its outputs. The LLaDA2 grids ran at buffer depth $2$ and the SDAR grids at buffer depth $4$, each on one engine build fixed within its family, and every grid compares configurations of one family, one build and one buffer depth. All configurations of the four main grids share block size $32$, admission threshold $0.10$, greedy decoding, a budget of $4{,}096$ total tokens, $4{,}096$ new tokens and $1{,}024$ forwards for each request, compiled execution with graph capture, and a batch cap of $8$ requests, except that the LLaDA2 code grid uses $16$ in all eight of its configurations.

Accept thresholds range over $\{0.85, 0.90, 0.95, 0.99\}$ and semi-completion thresholds over $\{0.70, 0.90\}$, with the engine's default, acc95/semi90, as the reference of each family. The LLaDA2 math grid holds seven of these configurations, without acc99/semi70. Each subtask contributes at most $512$ prompts, which binds only for GSM8K, so MATH-500 ($500$), HumanEval+ ($164$), MBPP ($500$) and MBPP+ ($378$) enter whole.

The SDAR grids need one engine change. The upstream engine fixes the SDAR accept threshold at $0.95$, and a small sampler-level change makes it configurable, with no training and no new weights. The default $0.95$ stays the reference of that family. The same build carries the instrumentation of the measurements of Section~\ref{sec:cmech}, disabled in every grid run.

\subsection{TPF and measured throughput}
\label{app:tpf-tps}

The serving logs of the grid configurations record total time together with end-to-end throughput. The LLaDA2 grids and the SDAR code grid each ran on one host, and the SDAR math grid ran on four, with five of its configurations on one host and one on each of the others. Every host of a family ran the same engine build with the settings fixed by the resolved configurations. On every host that served more than one configuration, TPF orders the configurations exactly as end-to-end throughput does, with Spearman $\rho = 1.0$, and the configuration with the highest TPF is also the fastest.

The reason is the static graph. The wall time of a forward varies across the configurations of a host by $0.8\%$ on the LLaDA2 math host, $3.0\%$ on the LLaDA2 code host, $2.5\%$ on the SDAR code host and $0.4\%$ on the SDAR math host, so throughput is TPF divided by a host-level constant, and TPF ordering is throughput ordering at fixed hardware.

\section{Positioning Against Related Work}
\label{app:relatedmatrix}

Table~\ref{tab:related} condenses the positioning of Section~\ref{sec:related} along four properties, a distribution-free finite-sample guarantee about deployed quality, error control over a space of configurations rather than one threshold, causal evidence for the operating point, and one statement that carries across cost measures. The nearest prior guarantee deserves a note. Theorem~1 of Fast-dLLM \citep{wu2025fastdllm} is a deterministic step-level inequality that holds once the model's confidence is trusted, so it guarantees neither that trust nor any configuration space and carries no $\delta$. Redline adds the statistical guarantee that the theorem lacks. Like Redline, CALM \citep{schuster2022confident} measures its risk against a reference, the full model, and for a $0/1$ risk its clipped risk difference coincides with our joint loss.

\begin{table}[t]
\caption{Positioning against related work. Guarantee is a distribution-free finite-sample statement about deployed quality, multiplicity is error control over a space of configurations, mechanism is causal evidence for the operating point, and generality is one procedure across cost measures.}
\label{tab:related}
\centering\scriptsize
\setlength{\tabcolsep}{4pt}
\begin{tabular}{@{}>{\raggedright\arraybackslash}p{0.27\linewidth}>{\raggedright\arraybackslash}p{0.18\linewidth}>{\raggedright\arraybackslash}p{0.16\linewidth}>{\raggedright\arraybackslash}p{0.14\linewidth}>{\raggedright\arraybackslash}p{0.168\linewidth}@{}}
\toprule
\bfseries\boldmath line of work & \bfseries\boldmath guarantee & \bfseries\boldmath multiplicity & \bfseries\boldmath mechanism & \bfseries\boldmath generality \\
\midrule
BD-LM families and serving engine (BD3-LM, SDAR, LLaDA2.0 and 2.1, MBD-LM) &
  \xmark{} mean accuracy & \xmark{} hand-picked points &
  stall documented, unexplained & \xmark \\
learned-frontier line (distillation, multi-block visibility, RL shaping, in-place revision, Fast-dLLM and v2, D2F, AdaBlock, dParallel, LoPA, d3LLM, DMax) &
  \xmark & \xmark & \xmark & one cost each \\
Fast-dLLM Theorem~1 & deterministic, conditional on trusting model confidence, no $\delta$ &
  \xmark & \xmark & one threshold \\
CALM (LTT on AR early exit) & \cmark{} finite-sample, relative to the full model & one threshold & \xmark &
  early exit only \\
LTT and conformal risk control (frameworks) & \cmark & \cmark{} procedure &
  \multicolumn{2}{l}{not a serving system} \\
lossy AR acceleration (speculative decoding, Medusa typical acceptance, PTQ) &
  \xmark{} supply the settings & \xmark & \xmark & one cost each \\
\midrule
\textbf{Redline (this work)} & \cmark{} distribution-free, finite-sample, family-wise $\delta$ &
  \cmark{} Holm over the serving grid & \cmark{} controlled measurements of all three levers &
  \cmark{} accept, schedule, speculative decoding, quantization \\
\bottomrule
\end{tabular}
\end{table}

\end{document}